\pdfoutput=1
\documentclass[11pt]{article}

\usepackage{acl}

\usepackage{times}
\usepackage{latexsym}
\usepackage{amsmath}
\usepackage[T1]{fontenc}
\usepackage{booktabs}

\usepackage[utf8]{inputenc}

\usepackage{microtype}

\usepackage{inconsolata}

\usepackage{graphicx}

\usepackage[table]{xcolor}
\title{Improving the OOD Performance of Closed-Source LLMs on NLI \\Through Strategic Data Selection}

 \author{Joe Stacey\textsuperscript{1,2}, Lisa Alazraki\textsuperscript{1}, Aran Ubhi\textsuperscript{1}, Beyza Ermis\textsuperscript{3}, Aaron Mueller\textsuperscript{4}, Marek Rei\textsuperscript{1}\\ %\textbf{} \\
        \textsuperscript{1}Imperial College London, \textsuperscript{2} University of Sheffield, \textsuperscript{3}Cohere Labs,
        \textsuperscript{4}Boston University \\
         \texttt{j.stacey@sheffield.ac.uk} \\
        \texttt{\{lisa.alazraki20, marek.rei\}@imperial.ac.uk} \\
        \texttt{aran.ubhi@me.com, beyza@cohere.com, amueller@bu.edu} }

\begin{document}
\maketitle
\begin{abstract}
We investigate the robustness of fine-tuned Large Language Models (LLMs) for the task of Natural Language Inference (NLI), finding that the in-distribution gains from fine-tuning correspond to a large drop in out-of-distribution (OOD) performance. Despite the widespread use of closed-source LLMs, there are no robustness mitigation methods that work under their API fine-tuning constraints.
Existing methods to improve robustness typically require changing the fine-tuning process or large-scale data augmentation, methods that are infeasible or cost prohibitive for closed-source models. To address this, we propose strategically selecting the NLI fine-tuning data, prioritising more complex examples or replacing existing training examples with LLM-generated data. 
Prioritising more complex training examples improves performance on challenging OOD NLI datasets, while training with synthetic data leads to substantial improvements on easier OOD datasets. We find that synthetic examples are often too simple, and by prompting LLMs to create more complex synthetic data we can improve performance on both easy and challenging OOD datasets. 
Finally, we show that recent autoregressive LLMs are substantially more robust to distributional shifts compared to encoder models, and should be a preferred baseline for future research.\footnote{Project code: \url{https://github.com/joestacey/LLM_robustness_NLI}}

\end{abstract}

\section{Introduction}

Large Language Models (LLMs) now perform impressively across a range of natural language understanding tasks \cite{openai2024gpt4technicalreport, geminiteam2024geminifamilyhighlycapable, cohere2025commandaenterprisereadylarge}, with models learning from in-context examples in the prompt \cite{DBLP:conf/nips/BrownMRSKDNSSAA20}. While fine-tuning LLMs often results in further in-distribution improvements \cite{DBLP:journals/jocss/AlizadehKSDZBKG25, luo-etal-2025-semi, wang-etal-2025-data-whisperer}, it is less clear what effect fine-tuning has on out-of-distribution (OOD) performance. 
We investigate this effect, finding that fine-tuning often causes large OOD performance drops alongside the in-distribution performance gains. 
Surprisingly, despite the recent popularity of closed-source LLMs, there are no existing methods to mitigate the loss in robustness\footnote{We use the term robustness to refer to the model performance on unseen OOD test sets.} for these models.
We aim to improve the OOD performance of closed-source LLMs on the task of Natural Language Inference (NLI), following a large body of prior work improving model robustness on this task (see Appendix \ref{sec:extra_related_work}).

Improving the OOD performance of fine-tuned models has been extensively studied for smaller-scale open-source models \cite{ravichander-etal-2023-bias, karimi-mahabadi-etal-2020-end, mccoy-etal-2019-right, clark-etal-2019-dont, he-etal-2019-unlearn, belinkov-etal-2019-dont, gururangan-etal-2018-annotation, poliak-etal-2018-hypothesis}, but these methods are not yet compatible with closed-source LLMs which are often state-of-the-art.
In particular, closed-source models do not allow for any modification of the training process, and training closed-source models with more data is more expensive, making large-scale data augmentation cost prohibitive.
To address these limitations, we investigate data-centric strategies for fine-tuning, using a fixed training budget of 10,000 instances. Our approach involves strategically selecting which examples to use for fine-tuning, with the aim of mitigating the loss in OOD performance while keeping the in-distribution gains.

The data selection methods we investigate involve: 1) selecting more challenging annotated examples in the training set, and 2) leveraging the few-shot capabilities of LLMs to generate new training instances from different domains. We investigate the effectiveness of replacing small quantities of training data with synthetic data to maintain a fixed training budget, contrasting with previous work which relies on costly large-scale augmentation \cite{hosseini-etal-2024-synthetic, DBLP:journals/corr/abs-2412-09263, wang-etal-2023-ibadr, chen-etal-2023-disco, wu-etal-2022-generating, liu-etal-2022-wanli}. 
To the best of our knowledge, this is the first work to investigate methods to improve robustness which are applicable for closed-source models within a realistic training budget.

Our contributions are as follows: 
\begin{enumerate}
\itemsep-0.2em
    \item We show that fine-tuning autoregressive LLMs leads to substantial performance improvements in-distribution. However, this comes at the cost of significantly lower performance on challenging OOD datasets (Section \ref{sec:fewshot}).
    \item We demonstrate that LLMs consistently outperform encoder models on OOD benchmarks for NLI, even when using less than 2\% of the training data. We suggest that encoder models are no longer appropriate baselines, despite their continued use (Section \ref{sec:fewshot}).
    \item We propose strategies to increase the representation of challenging examples in the training data. This improves robustness on complex OOD datasets while retaining performance on less complex OOD data (Section \ref{sec:sampling_results}).
     \item We investigate training with LLM-generated synthetic data, finding that replacing a subset of the training data with LLM-generated examples leads to large improvements on less complex OOD data.
     Creating more complex synthetic data leads to further improvements, with better performance on both challenging and less challenging OOD data (Section \ref{sec:data_generation_results}).
\end{enumerate}

\section{Related Work}
There is limited and sometimes conflicting evidence about the effect of fine-tuning on OOD performance.
%compared to few-shot prompting. 
\citet{mosbach-etal-2023-shot} compare models fine-tuned with just 16 examples to models that use in-context learning with 16 examples in the prompt, finding that fine-tuning leads to better OOD performance for larger models. In contrast, contemporaneous work by \citet{lampinen2025generalizationlanguagemodelsincontext} finds that fine-tuning is less robust than in-context learning on reversals \cite{DBLP:conf/iclr/BerglundTKBSKE24} and other rule-based generalisations. Rather than considering specific rule-based generalisations, we evaluate performance across a wide range of different out-of-distribution NLI benchmarks.

%As rule-based evaluations are often overemphasised in existing work \cite{weissweiler2025linguisticgeneralizationsrulesimpacts}, 
%We instead assess robustness by evaluating OOD performance across a wide range of NLI benchmarks. 

For NLI, OOD performance is often improved by debiasing models to mitigate either known \cite{karimi-mahabadi-etal-2020-end, utama-etal-2020-mind, clark-etal-2019-dont, he-etal-2019-unlearn} or unknown \cite{utama-etal-2020-towards, cheng-amiri-2024-fairflow, clark-etal-2020-learning, DBLP:conf/iclr/Sanh0BR21} dataset biases.
This typically involves weighting the loss of more biased examples \cite{karimi-mahabadi-etal-2020-end, clark-etal-2019-dont, clark-etal-2020-learning}, or incorporating soft predictions from an intentionally biased model during training \cite{karimi-mahabadi-etal-2020-end, utama-etal-2020-mind}.
However, debiasing against one type of bias can inadvertently increase a model’s reliance on other biases \cite{ravichander-etal-2023-bias}. %Moreover, these debiasing techniques are not applicable to closed-source LLMs, where training is only accessible through an API and the training process cannot be modified.
Alternatively, OOD performance can be improved by augmenting with additional minority examples\footnote{Minority examples are those that counter frequent spurious patterns in the dataset \cite{tu-etal-2020-empirical}} during training, which are identified via model misclassifications on the training data \cite{DBLP:conf/icml/LiuHCRKSLF21}, high variance across pruned subnetworks \cite{du-etal-2023-robustness}, or label flips during training \cite{yaghoobzadeh-etal-2021-increasing}. %Alternatively, \citet{korakakis-vlachos-2023-improving} use a minimax objective to upweight loss on more challenging examples. 
Inspired by these methods, we aim to make more challenging examples better represented when training closed-source LLMs. %In contrast to previous work, we do this while maintaining the size of the training data.

\begin{figure*}
\begin{center}
    \includegraphics[width=380pt]{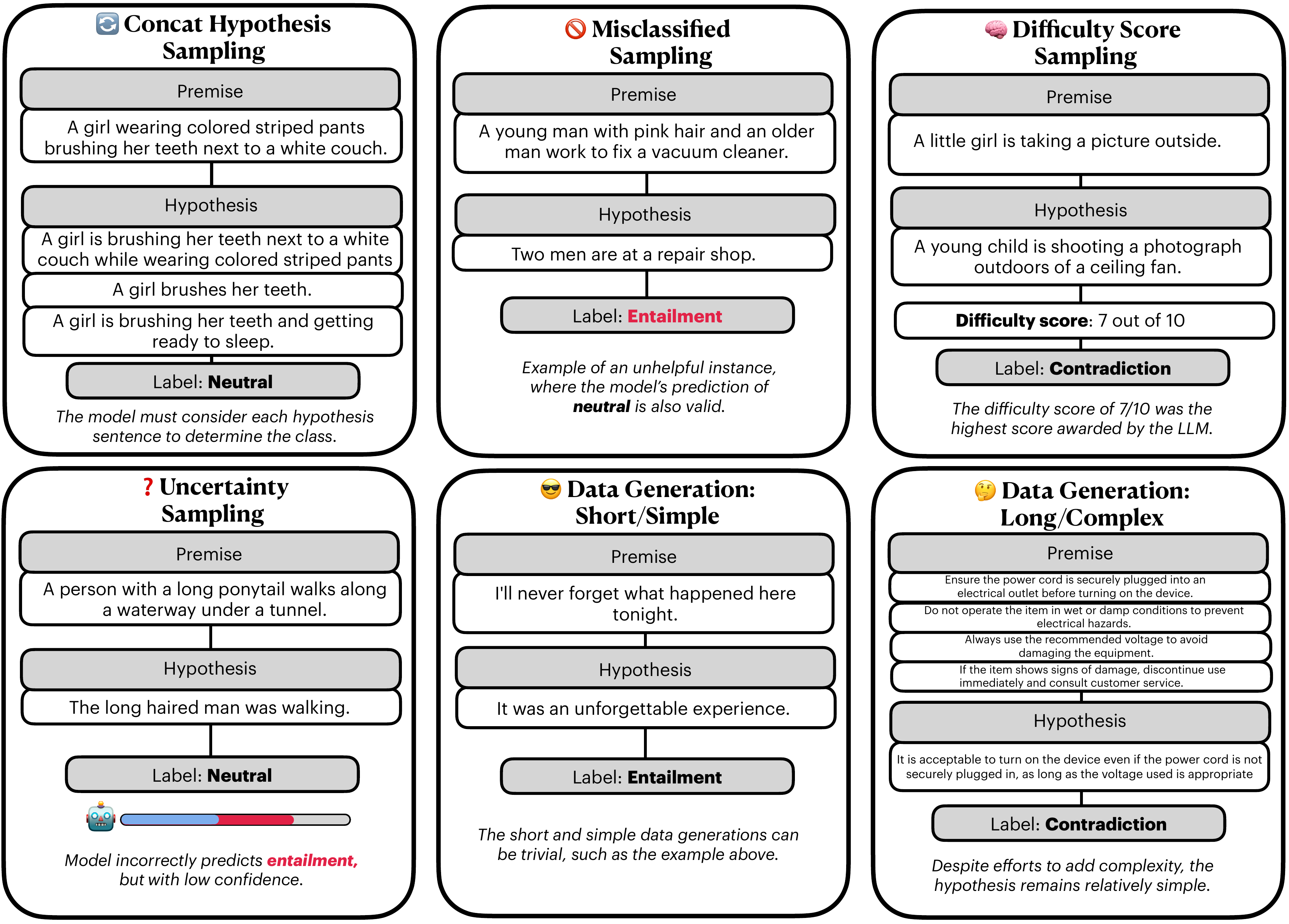} 
    \caption{Examples of a training instance in $\mathcal{D}_\text{up}$ from our different methods.} 
    \label{fact_gen_diagram}

\end{center}
\end{figure*}

The rise of LLMs has made data augmentation a popular strategy for improving performance and robustness \cite{wu-etal-2022-generating, liu-etal-2022-wanli, wang-etal-2023-ibadr, chen-etal-2023-disco, hosseini-etal-2024-synthetic, DBLP:journals/corr/abs-2412-09263}. However, despite the recent rapid advances in LLMs, few-shot labelling for NLI remains a challenging task \cite{hosseini-etal-2024-synthetic, DBLP:journals/corr/abs-2311-09825}. This limitation is sometimes addressed through additional fine-tuning of the generation model \cite{hosseini-etal-2024-synthetic}, human annotation \cite{liu-etal-2022-wanli, he-etal-2023-targeted}, teacher models that estimate label distributions \cite{stacey-rei-2024-distilling}, or applying gradient surgery \cite{DBLP:journals/corr/abs-2410-10865}. 
Such large-scale data augmentations are inefficient, and are often cost prohibitive when fine-tuning closed-source LLMs. Therefore, in this work we instead consider a fixed training budget, where we replace existing human-annotated instances with LLM-generated synthetic data. 
Prior work finds that replacing human‑annotated data with synthetic data can reduce in‑distribution performance \cite{ashok2024little}, but without considering OOD performance which is the main focus of this work.

For a more comprehensive review of NLI robustness work, see Appendix~\ref{sec:extra_related_work}. 

\section{Methods} \label{sec:methods}

We aim to improve the robustness of a closed-source LLM, $\mathcal{M}$, under a fixed training budget by: 1) increasing the proportion of challenging training examples, and 2) replacing a subset of the original training data with LLM-generated data from different domains. Our experiments show that targeted replacement is sufficient to achieve strong improvements in OOD robustness, and that replacing fewer training examples can lead to larger improvements (see Appendix~\ref{appendix_subsec:ablation_experiments}).  Partial replacement also minimises inference costs when generating the synthetic data. 

\subsection{Definitions}
Let $\mathcal{D} = \{(x_1, y_1), \ldots, (x_n, y_n)\}$ denote a large NLI dataset, where each $x_i$ is a premise–hypothesis pair and $y_i \in \{\text{entailment, neutral, contradiction}\}$. We define an initial training subset $\mathcal{D}_\text{init} \subset \mathcal{D}$ of size $m$, which is used to fine-tune $\mathcal{M}$, resulting in a baseline model $\mathcal{M}_{\text{base}}$.
We then construct two additional subsets: $\mathcal{D}_{\text{up}}$, representing new, challenging examples to include in the training set, and $\mathcal{D}_{\text{down}} \subset \mathcal{D}_{\text{init}}$, representing existing examples to remove such that the final training set $(\mathcal{D}_{\text{init}} \cup \mathcal{D}_{\text{up}}) \setminus \mathcal{D}_{\text{down}}$ maintains size $m$. 
To avoid confounding effects from shifts in the label distribution,  we ensure $|\mathcal{D}_{\text{up}}^c| = |\mathcal{D}_{\text{down}}^c| \leq \mathcal{K}$ for each label $c$, where $\mathcal{D}_{\text{up}}^c$ denotes the examples in $\mathcal{D}_{\text{up}}$ with label $c$.
When inference is required to select examples from outside $\mathcal{D}_{\text{init}}$, we define a candidate pool $\mathcal{D}_{\text{potential}} \subset \mathcal{D} \setminus \mathcal{D}_{\text{init}}$ with $|\mathcal{D}_{\text{potential}}| = m$ to reduce cost. While inference is cheaper than fine-tuning, performing inference over the full dataset is still computationally expensive.  

We now describe a range of different methods for constructing $\mathcal{D}_{\text{up}}$, which either involve selecting challenging training examples, or using LLMs to generate new synthetic examples.% and $\mathcal{D}_{\text{down}}$.

\subsection{Uncertainty Sampling}

The confidence of model predictions can suggest how challenging an instance is, with high confidence in the correct class suggesting a lack of difficulty, while high confidence in the wrong class may instead suggest annotation errors \cite{swayamdipta-etal-2020-dataset}. We therefore choose examples based on maximising the uncertainty of the model predictions, choosing $\mathcal{D}_\text{up}^\text{c}$ as the top $\mathcal{K}$ examples in $\mathcal{D}_\text{potential}^\text{c}$ with the highest entropy over the soft predictions from $\mathcal{M}_\text{base}$. This assumes the availability of the model output probabilities.\footnote{These output probabilities are available for GPT-4o-mini, but not the closed-source Gemini and Command R models that we experiment with} Unlike \citet{swayamdipta-etal-2020-dataset, liu-etal-2022-wanli}, we are unable to identify any changes in predictions during training due to the closed-source nature of the models. 

\subsection{Difficulty Score Sampling}

We aim to exploit the wide-ranging capabilities of few-shot LLMs, using the models to assess the difficulty of each instance in the training set, before using this information to help improve the robustness of the fine-tuned LLM. To achieve this, we prompt $\mathcal{M}$ to assess the difficulty of each labelled instance in $\mathcal{D}_\text{potential}$, providing a score from 1 to 10, before choosing the top $\mathcal{K}$ scored examples from $\mathcal{D}_\text{potential}^\text{c}$. We also experiment with finding scores for the label correctness, plausibility and fluency of each example.

\subsection{Misclassified Sampling}

Inspired by prior work upsampling minority examples \cite{DBLP:conf/icml/LiuHCRKSLF21, yaghoobzadeh-etal-2021-increasing}, we use our baseline model $\mathcal{M}_\text{base}$ to make predictions on $\mathcal{D}_\text{potential}$, choosing $\mathcal{K}$ examples for each class that were misclassified. As fewer than $\mathcal{K}$ examples may be misclassified for a particular class, we have $|\mathcal{D}_\text{up}^\text{c}| = |\mathcal{D}_\text{down}^\text{c}| \leq \mathcal{K}$. While the resulting training data may contain examples with incorrect labels, the average difficulty of the training sample is also likely to be greater.

\subsection{Concatenative Hypothesis Sampling}

Rather than selecting $\mathcal{D}_\text{up}$ as a subset of $\mathcal{D} \setminus \mathcal{D}_\text{init}$, we experiment with deriving more complex instances using the examples in $\mathcal{D} \setminus \mathcal{D}_\text{init}$.\footnote{As no inference is required, we do not need to restrict this method to using $\mathcal{D}_\text{potential}$} Specifically, we identify instances with the same premise,\footnote{It is common for large-scale NLI datasets to involve instances with repeated premises \cite{bowman-etal-2015-large, williams-etal-2018-broad, nie-etal-2020-adversarial}} and concatenate their corresponding hypotheses to create more challenging examples. 

%To decide the class of these new, more challenging instances: 
To assign a label to these new instances, we use the following simple rules: 
if any single hypothesis is a contradiction, then the combined hypothesis is a contradiction. Otherwise, if any single hypothesis is neutral, then the combined hypothesis is neutral. If there is no contradiction or neutral hypothesis, then the combined hypothesis must be entailment. This strategy allows us to create more challenging instances out of the existing NLI training data. We choose $\mathcal{D}_\text{up}^\text{c}$ by randomly selecting $\mathcal{K}$ instances that concatenate $\mathcal{H}$ hypotheses, where the concatenated hypotheses belong to class c.

\subsection{Few-Shot LLM Data Generation}

We generate additional, synthetic data using $\mathcal{M}$, our LLM before the fine-tuning, without relying on any external models. This data is produced in a zero-shot setting across a range of domains (see Appendix~\ref{appendix_sec:further_details}), with each instance containing a single-sentence premise and hypothesis (Short \& Simple Generation). We also use $\mathcal{M}$ to label unlabelled data generated for the MNLI \cite{williams-etal-2018-broad} training domains provided by \citet{stacey-rei-2024-distilling}\footnote{See Appendix \ref{appendix_sec:further_details} for the additional filtering we do on this data to improve the quality} (MNLI Domains Generation), which was generated by an older \texttt{text-curie-001} GPT-3 model. To increase data complexity, we prompt the LLM to generate a new four-sentence premise (Long \& Simple Generation). To further raise the difficulty, we also prompt the LLM with guidance about how the hypothesis should relate to multiple parts of the premise, resulting in more contextually dependent examples (Long \& Complex Generation - see Appendix \ref{appendix_sec:further_details} for details).

\definecolor{light-gray}{gray}{0.94}
\begin{table}
\centering
\resizebox{\columnwidth}{!}{
 \begin{tabular}{l c c c}
  \toprule
    & SNLI (ID) & Challenge-OOD & Standard-OOD \\
  \midrule
 \multicolumn{4}{l}{\textbf{GPT-4o-mini:}} \\
Without fine-tuning & 85.02 \text{ } & 64.75 \text{ }  & 83.03 \text{ }  \\
With fine-tuning (10k) & 92.47$\uparrow$ & 59.62$\downarrow$ & 79.63$\downarrow$ \\
  \midrule
    \multicolumn{4}{l}{\textbf{Gemini-2.0-Flash:}} \\
Without fine-tuning & 82.39 \text{ } & 76.85 \text{ } & 79.15 \text{ } \\
With fine-tuning (10k) & 92.61$\uparrow$ & 65.90$\downarrow$ & 81.38$\uparrow$ \\
\bottomrule
  \end{tabular}}
\caption{Accuracy of both GPT-4o-mini and Gemini-2.0-Flash both before and after fine-tuning with 10,000 training instances. Fine-tuning substantially improves in-distribution (ID) performance on SNLI, but at the expense of OOD performance on challenging NLI datasets. See Appendix \ref{appendix_subsec:baseline_analysis} for the full results for each dataset.}
\label{tab:LLMs_with_and_without_FT}
\end{table}

\definecolor{light-gray}{gray}{0.94}
\begin{table*}
\centering
\resizebox{\textwidth}{!}{
 \begin{tabular}{l c c c c c c c c @{\hskip 0.5em} c @{\hskip 0.5em} c @ {\hskip 0.5em} 
 c c c c}
  \toprule

& & \multicolumn{6}{c}{Challenge-OOD} & & \multicolumn{5}{c}{Standard-OOD}  \\ 
    \cmidrule{3-8} \cmidrule{10-14} 
    & SNLI & r1 & r2 & r3 \text{ } & COPA & INLI-I & WANLI & Avg. & MNLI-m & MNLI-mm & FEVER & Scitail & INLI-NLI & Avg. \\
  \midrule
    %\multicolumn{15}{c}{\textbf{Encoder models (full training data)}} \\
    \rowcolor{light-gray}\multicolumn{15}{c}{Encoder models (full training data)} \\
RoBERTa-large & 92.63 \text{   } \text{   }  & 42.90 & 29.36 & 28.28 & 51.12 & 47.72 & 56.32 & 42.62 \text{   } \text{   }  & 85.82 & 85.26 & 66.41 & 71.07 & 76.04 & 76.92 \text{   } \text{   }  \\% & 29.36 & 28.28 & 51.12 & 47.72 & 56.32 & 42.62 \\% & 66.41 & 85.82 & 85.26 & 71.07 & 76.04 & 76.92 \\
DeBERTa-base & 92.64 \text{   } \text{   }  & 40.34 & 34.08 & 31.82 & 50.92 & 36.88 & 57.91 & 41.99 \text{   } \text{   }  & 84.68 & 85.25 & 63.68 & 75.82 & 72.17 & 76.32 \text{   } \text{   }   \\ % & 34.08 & 31.82 & 50.92 & 36.88 & 57.91 & 41.99 \\% & 63.38 & 84.68 & 85.25 & 75.82 & 72.17 & 76.26 \\
DeBERTa-large & \textbf{93.13} \text{   } \text{   }  & 53.15 & 41.53 & 37.13 & 51.38 & 45.80 & 61.55 & 48.42 \text{   } \text{   }  & 87.75 & 88.02 & 68.42 & 74.93 & 78.49 & 79.52 \text{   } \text{   }  \\ %& 41.53 & 37.13 & 51.38 & 45.80 & 61.55 & 48.42 \\% & 68.42 & 87.75 & 88.02 & 74.93 & 78.49 & 79.52 \\
  \midrule
%    \multicolumn{15}{c}{\textbf{Debiasing methods w/ DeBERTa-large:}} \\
%    JTT & 90.86$\downarrow$$\dagger$ & 51.40 & 40.65 & 35.92 & 53.33 & 43.08 & 61.68 & 47.67$\downarrow$ \text{  }  & 85.47 & 85.70 & 67.06 & 74.09 & 77.52 & 77.97$\downarrow$ \text{  }  \\% & 40.65 & 35.92 & 53.33 & 43.08 & 61.68 & 47.68$\uparrow$ \\%& 67.07 & 85.47 & 85.70 & 74.09 & 77.52 & 77.97$\downarrow$ \\
%    PoE & 87.67$\downarrow$$\dagger$ & 47.58 & 40.65 & 40.17 & 50.53 & \textbf{89.80} & 55.53 & 54.04$\uparrow$$\dagger$& 79.43 & 79.21 & 62.43 & 81.60 & 69.68 & 74.47$\downarrow$$\dagger$  \\%& 40.65 & 40.17 & 50.53 & \textbf{89.80} & 55.53 & 54.04$\uparrow$ \\% & 62.43 & 79.43 & 79.21 & 81.60 & 69.68 & 74.47$\downarrow$ \\
%    Reweight & 91.78 $\downarrow$\text{  } & 52.35 & 41.80 & 40.33 & 50.30 & 67.55 & 59.56 & 51.98$\uparrow$\text{  } & 86.36 & 85.67 & 68.00 & 81.75 & 77.18 & 79.79$\uparrow$\text{  } \\%& 41.80 & 40.33 & 50.30 & 67.55 & 59.56 & 51.98$\uparrow$ \\ %& 68.00 & 86.36 & 85.67 & 81.75 & 77.18 & 79.79$\uparrow$  \\
%    \midrule
    %\multicolumn{15}{c}{\textbf{LLMs (with 10,000 training instances)}} \\
    \rowcolor{light-gray}\multicolumn{15}{c}{LLMs (with 10,000 training instances)} \\
    
Command R & 91.14$\downarrow$ \text{  } & 63.58 & 55.78 & 52.88 & 56.60 & \textbf{66.84} & 61.21 & 59.48$\uparrow$$\dagger$ & 86.99 & 87.02 & 70.37 & \textbf{85.84} & 82.35 & \textbf{82.51}$\uparrow$$\dagger$ \\%& 55.78 & 52.88 & 56.60 & 66.84 & 61.21 & 59.48$\uparrow$ \\% & 58.04 & 86.99 & 87.02 & \textbf{85.84} & 82.35 & 80.05$\uparrow$ \\
Gemini-2.0-Flash & 92.61$\downarrow$$\dagger$ & \textbf{72.68} & \textbf{64.10} & \textbf{61.45} & \textbf{75.76} & 58.82 & \textbf{62.61} & \textbf{65.90}$\uparrow$$\dagger$ & \textbf{88.30} & \textbf{88.54} & \textbf{71.56} & 73.42 & \textbf{85.09} & 81.38$\uparrow$$\dagger$ \\% & \textbf{64.10} & \textbf{61.45} & \textbf{75.76} & 58.82 & \textbf{62.61} & \textbf{65.90}$\uparrow$ \\%& \textbf{71.56} & \textbf{88.30} & \textbf{88.54} & 73.42 & \textbf{85.09} & \textbf{81.38}$\uparrow$ \\
GPT-4o-mini & 92.47$\downarrow$$\dagger$ & 65.98 & 58.10 & 55.38 & 56.20 & 61.42 & 60.62 & 59.62$\uparrow$$\dagger$ & 86.93 & 87.24 & 71.21 & 72.55 &  80.21 & 79.63$\uparrow$ \text{  } \\% & 58.10 & 55.38 & 56.20 & 61.42 & 60.62 & 59.62$\uparrow$ \\%& 71.21 & 86.93 & 87.24 & 72.55 &  80.21 & 79.63$\uparrow$ \\
%\midrule
%\multicolumn{15}{c}{\textbf{LLMs (with no fine-tuning):}} \\
%Gemini-2.0-Flash & 82.39 \text{   } %\text{   } & 79.70 & 71.50 & 67.08 & 92.30 & 84.60 & 65.42 & 76.85 \text{   } \text{   } & 83.24 & 82.45 & 72.9 & 72.58 & 84.60 & 79.15 \text{   } \text{   } \\
%GPT-4o-mini & 85.02 \text{   } \text{   } & 77.00 & 63.90 & 58.33 & 63.60 & 60.80 & 64.88 & 64.75 \text{   } \text{   }& 84.62 & 85.09 & 73.53 & 84.62 & 87.30 & 83.03 \text{   } \text{   } \\
\bottomrule
  \end{tabular}}
\caption{The performance of RoBERTa \cite{liu2019robertarobustlyoptimizedbert} and DeBERTa \cite{DBLP:conf/iclr/HeLGC21, DBLP:conf/iclr/HeGC23} models fine-tuned on the full SNLI training set are compared to Command R, Gemini-2.0-Flash and GPT-4o-mini models fine-tuned with just 10,000 instances. Accuracy is tested across a range our challenging NLI test sets (Challenge-OOD) and our other NLI test sets (Standard-OOD). The best results from fine-tuned models are in \textbf{bold}. $\uparrow$ and $\downarrow$ show better or worse average performance compared to the DeBERTa-large baseline. $\dagger$ represents results where $p < 0.05$, using two-tailed bootstrapping statistical testing \cite{efron1993introduction}}
\label{tab:Encoders_vs_LLMs}
\end{table*}

Finally, we investigate whether label quality can be improved, given that LLMs are known to be unreliable annotators. Models trained on LLM-generated labels can underperform compared to those trained on human annotated data \cite{DBLP:conf/icbinb/MohtaAXS23}, with previous work cautioning against fully relying on LLMs for annotation \cite{DBLP:journals/corr/abs-2408-05534, brassard-etal-2022-copa}. While few-shot LLM annotation has shown promise for some tasks \cite{DBLP:journals/corr/abs-2501-10970, DBLP:journals/corr/abs-2303-15056, DBLP:journals/corr/abs-2304-06588}, it remains unreliable for NLI \cite{DBLP:journals/corr/abs-2311-09825, hosseini-etal-2024-synthetic}. Inspired by the \textit{if in doubt, discard} approach from task-oriented dialogue \cite{stacey-etal-2024-lucid}, we generate eight few-shot predictions from $\mathcal{M}$ per instance, and retain only those for which all predictions agree.

\section{Experiments}

We evaluate $\mathcal{M}$, $\mathcal{M}_\text{base}$, and each proposed method on a diverse range of out-of-distribution NLI datasets: WANLI \cite{liu-etal-2022-wanli}, ANLI (r1, r2 and r3) \cite{nie-etal-2020-adversarial}, Scitail \cite{Khot_Sabharwal_Clark_2018}, MNLI \cite{williams-etal-2018-broad}, and INLI (including the implied entailment subset, INLI-I, and the remainder, which we describe as INLI-NLI) \cite{havaldar2025entailedlinesincorporatingimplication}. We further include FEVER-NLI \cite{thorne-etal-2018-fever, DBLP:conf/aaai/NieCB19}, a fact verification dataset converted to NLI, and introduce COPA-NLI, a challenging NLI test set constructed from Balanced-COPA \cite{kavumba-etal-2019-choosing, gordon-etal-2012-semeval} (see Appendix~\ref{appendix_sec:experimental_details}). %For INLI, we report separate results for the challenging \textit{implied entailment} subset (INLI-I), and the remaining subset (INLI-NLI). 

We categorise each NLI dataset as either Challenge-OOD or Standard-OOD, defining Challenge-OOD as any dataset where the GPT-4o-mini baseline achieves below 70\% accuracy. Using this threshold, we find that the resulting Challenge-OOD datasets are either intentionally designed to be challenging \cite{nie-etal-2020-adversarial, havaldar2025entailedlinesincorporatingimplication} or are constructed using more difficult training examples \cite{liu-etal-2022-wanli}.

We test the robustness of three models available for fine-tuning as closed-source models: GPT-4o-mini (gpt-4o-mini-2024-07-18),\footnote{https://platform.openai.com/docs/models/gpt-4o-mini} Command R (base\_type\_chat),\footnote{https://cohere.com/blog/commandr-fine-tuning} and Gemini-2.0-flash (gemini-2.0-flash-001).\footnote{https://cloud.google.com/vertex-ai/generative-ai/docs/models/gemini/2-0-flash} Unless stated otherwise, our experiments use GPT-4o-mini as a baseline. To mitigate the high variance in OOD predictions \cite{mccoy-etal-2020-berts}, each result is an average from fine-tuning five models using different random seeds.

Each of the proposed methods assume no access to OOD data for hyper-parameter tuning.
%While tuning hyper-parameters on OOD data has been found to lead to strong robustness gains, we argue that this is a less realistic robustness setting. 
In our experiments, we set $\mathcal{K}$ to 5\% of the training sample size, and fix $m$ at 10,000 training instances. We set $\mathcal{H}$ to be 3 for our Concatenative Hypothesis Sampling. % We could justify here about why m=10,000 is a good choice.
When training the encoder models, we train the models with the full SNLI training data, in line with common practice. See Appendix \ref{appendix_sec:experimental_details} for more detail about hyper-parameter choices. %
As a baseline, we also include \textit{Random Sampling}, where we choose $\mathcal{D}_\text{up}^\text{c}$ as $\mathcal{K}$ randomly selected examples from $\mathcal{D}_\text{potential}^\text{c}$. We use SNLI \cite{bowman-etal-2015-large} as our training data, with additional experiments using MNLI training data in Appendix \ref{appendix_subsec:mnli_experimentation}.  

Finally, while some of the methods for identifying examples in $\mathcal{D}_\text{up}$ can also be used to identify examples for $\mathcal{D}_\text{down}$, this is not possible for all methods. Therefore, for comparability across methods, we choose $\mathcal{D}_\text{down}$ by random sampling from $\mathcal{D}_\text{init}$, avoiding making further changes to the training distribution. In Appendix~\ref{appendix_subsec:ablation_experiments}, we explore the effect of using alternative strategies for $\mathcal{D}_{\text{down}}$.

%Our results are supported by extensive experimentation, where we fine-tune over 200 models, evaluating each model on 12 different test-sets.
%
In total, we fine-tune 200+ models and evaluate each on 12 datasets, totalling 2,400+ different model-dataset evaluations.

\begin{table*}
\centering
\resizebox{\textwidth}{!}{
 \begin{tabular}{l c c c c c c c c @{\hskip 0.5em} c @{\hskip 0.5em} c @ {\hskip 0.5em} 
 c c c c}
  \toprule

& & \multicolumn{6}{c}{Challenge-OOD} & & \multicolumn{5}{c}{Standard-OOD}  \\ 
    \cmidrule{3-8} \cmidrule{10-14} 
    & SNLI & r1 & r2 & r3 \text{ } & COPA & INLI-I & WANLI & Avg. & MNLI-m & MNLI-mm & FEVER & Scitail & INLI-NLI & Avg. \\
  \midrule
%  \multicolumn{15}{c}
%{\textbf{LLM with no FT:}} \\
%LLM w/o FT & 85.02 \text{   } \text{   } & 77.00 & 63.90 & 58.33 & 63.60 & 60.80 & 64.88 & 64.75 \text{   } & 84.62 & 85.09 & 73.53 & 84.62 & 87.30 & 83.03 \text{   } \text{   } \\
%\midrule

\rowcolor{light-gray}\multicolumn{15}{c}{Fine-tuning Baselines} \\
Baseline (10k) & 92.47 \text{   } \text{   } & 65.98 & 58.10 & 55.38 & 56.20 & 61.42 & 60.62 & 59.62 \text{  } \text{    }  & 86.93 & 87.24 & 71.21 & 72.55 & 80.21 & 79.63 \text{  } \text{   } \\% & 65.98 & 58.10 & 55.38 & 56.20 & 61.42 & 60.62 & 59.62 \text{ } \\ %& 71.21 & 86.93 & 87.24 & 72.55 & 80.21 & 79.63 \text{  }\\
% &  &  &  &  &  &  &  & - & & &  & & & - \\
Random Sampling & 92.55$\uparrow$ \text{   } & 65.80 & 58.66 & 55.37 & 56.48 & 60.60 & 60.15 & 59.51$\downarrow$ \text{    } & 86.94 & 87.00 & 71.17 & 69.56 & 81.15 & 79.16 $\downarrow$ \text{   } \\ % & 65.80 & 58.66 & 55.37 & 56.48 & 60.60 & 60.15 & 59.51$\downarrow$ \\% & 71.17 & 86.94 & 87.00 & 69.56 & 81.15 & 79.16 $\downarrow$ \\
% & +0.08 & (0.18) & +0.56 & (0.01) & +0.28 & (0.82) & (0.46) & (0.11) & (0.04) & +0.01 & (0.24) & (2.99) & +0.94 & -0.46 \\
% &  &  &  &  &  &  &  & (0.11) & & &  & & & (0.46) \\
   \midrule

% & (0.15) & (0.50) & (0.12) & (2.65) & +6.75 & (4.54) & (0.30) & (0.23) & +0.64 & (0.06) & (0.18) & (5.03) & (0.41) & (1.01) \\
% &  & &  &  & &  & & (0.23) & &  &  &  &  & (1.01) \\
  \rowcolor{light-gray} \multicolumn{15}{c}{Sampling Methods} \\
Misclassified Sampling & 92.32$\downarrow$\text{   }  & 65.48 & 57.98 & 52.73 & 62.94 & 56.88 & 60.32 & 59.39$\downarrow$ \text{    } & 86.87 & 87.06 & 71.85 & 67.52 & 79.80 & 78.62 $\downarrow$ \text{   } \\ %& 65.48 & 57.98 & 52.73 & 62.94 & 56.88 & 60.32 & 59.39$\downarrow$\\% & 71.85 & 86.87 & 87.06 & 67.52 & 79.80 & 78.62 $\downarrow$ \\

% Correct \& Uncertain & 92.68 & 66.90 & 57.52 & 54.88 & 57.10 & 59.06 & \textbf{61.62} & 59.51$\downarrow$ & 71.74 & 88.21 & 88.24 & 75.39 & 80.88 & 80.89 $\uparrow$\\
% &  &  &  &  &  &  &  & (0.10) & &  &  & & & +1.26 \\
% & +0.08 & +1.44 & +0.24 & (0.40) & +1.38 & (1.88) & +0.46 & +0.21 & +0.38 & +0.86 & +0.69 & (1.93) & +0.40 & +0.08 \\
% &  &  & &  &  &  &  & +0.21 & & & & & & +0.08 \\
Concat Hypothesis Sampling & 92.56$\uparrow$\text{   }  & 65.72 & 58.14 & 55.88 & 62.84 & \textbf{62.04} & 60.00 & 60.77$\uparrow$ \text{    } & 86.55 & 86.69 & 70.81 & 69.88 & 80.71 & 78.93 $\downarrow$ \text{   } \\ % & 65.72 & 58.14 & 55.88 & 62.84 & \textbf{62.04} & 60.00 & 60.77$\uparrow$ \\ % & 70.81 & 86.55 & 86.69 & 69.88 & 80.71 & 78.93 $\downarrow$\\
% & +0.09 & (0.26) & +0.04 & +0.50 & +6.64 & +0.02 & (0.62) & +1.05 & (0.40) & (0.38) & (0.55) & (2.67)  & +0.50 & (0.70) \\
% & &  & &  &  &  &  & +1.05 & &  &  &   & & (0.70) \\
Difficulty Score Sampling & 92.66$\uparrow$$\dagger$ & 67.48 & 59.10 & 56.42 & 58.84 & 59.54 & 61.40 & 60.46$\uparrow$ \text{    } & 87.64 & 87.66 & 71.61 & 71.88 & 80.81 & 79.92 $\uparrow$ \text{   } \\ % & 67.48 & 59.10 & 56.42 & 58.84 & 59.54 & 61.40 & 60.46$\uparrow$ \\ % & 71.61 & 87.64 & 87.66 & 71.88 & 80.81 & 79.92 $\uparrow$
Uncertainty Sampling & \textbf{92.80}$\uparrow$$\dagger$ & 67.42 & 58.60 & 55.12 & \textbf{63.68} & 60.60 & 60.99 & \textbf{61.07}$\uparrow$ \text{    } & \textbf{88.14} & 88.24  & 71.57 & 71.47 & 82.01 & 80.28 $\uparrow$ \text{   } \\% & 67.42 & 58.60 & 55.12 & \textbf{63.68} & 60.60 & 60.99 & \textbf{61.07}$\uparrow$ \\ % & 71.57 & \textbf{88.14} & 88.24 & 71.47 & 82.01 & 80.28 $\uparrow$ \\
% & +0.33 & +1.44 & +0.50 & (0.26) & +7.48 & (0.82) & +0.37 & +1.45 & +0.36 & +1.21 & +1.00 & (1.08) & +1.80 & +0.66 \\
% &  &  &  & &  &  &  & \textbf{+1.45} &  &  &  &  &  & +0.66 \\
 \midrule
   \rowcolor{light-gray} \multicolumn{15}{c}{Generated Data} \\

MNLI Domains Generation & 92.59$\uparrow$ \text{    }  & 67.72 & 57.96 & 54.00 & 54.80 & 56.04 & 62.89 & 58.90$\downarrow$ \text{    } & 87.79 & 88.34 & 72.97 & 80.98 & \textbf{85.90} & 83.20 $\uparrow$$\dagger$ \\ % & 67.72 & 57.96 & 54.00 & 54.80 & 56.04 & 62.89 & 58.90$\downarrow$ \\ % & 72.97 & 87.79 & 88.34 & 80.98 & 85.90 & 83.20 $\uparrow$ \\

Short \& Simple Generation & 92.47 \text{   }  \text{    }  & 68.88 & 58.14 & 53.43 & 56.28 & 52.02 & \textbf{64.11} & 58.81$\downarrow$ \text{    } & 87.88 & 88.66 & \textbf{73.48} & \textbf{81.67} & 84.76 & \textbf{83.29} $\uparrow$$\dagger$ \\ %& 68.88 & 58.14 & 53.43 & 56.28 & 52.02 & \textbf{64.11} & 58.81$\downarrow$ \\ % & 73.48 & 87.88 & 88.66 & \textbf{81.67} & 84.76 & \textbf{83.29} $\uparrow$ \\

Long \& Simple Generation & 92.53$\uparrow$ \text{    } & \textbf{69.62} & \textbf{60.40} & 52.58 & 55.22 & 50.00 & 62.75 & 58.43$\downarrow$ \text{    } & 88.08 & \textbf{88.88} & 73.05 & 78.97 & 83.33 & 82.46 $\uparrow$$\dagger$ \\ %& \textbf{69.62} & 60.40 & 52.58 & 55.22 & 50.00 & 62.75 & 58.43$\downarrow$ \\ %& 73.05 & 88.08 & \textbf{88.88} & 78.97 & 83.33 & 82.46 $\uparrow$ \\

Long \& Complex Generation & 92.26$\downarrow$ \text{    } & 69.34 & 60.28 & \textbf{57.42} & 55.90 & 61.42 & 60.86 & 60.87$\uparrow$$\dagger$ & 87.74 & 88.07 & 72.85 & 76.72 & 85.83 & 82.24 $\uparrow$$\dagger$\\ %& 67.24 & \textbf{59.04} & \textbf{56.03} & 55.68 & 61.14 & 60.35 & 59.91$\uparrow$ \\ %& \textbf{72.06} & 87.56 & 88.03 & 76.20 & \textbf{85.73} & 81.92 $\uparrow$ \\
% & +0.17 & +1.26 & +0.94 & +0.65 & (0.52) & (0.28) & (0.27) & +0.30 & +0.85 & +0.63 & +0.79 & +3.65 & +5.52 & +2.29 
% &  &  &  &  &  &  &  & +0.30 &  &  &  &  & & \textbf{+2.29} 
\bottomrule
  \end{tabular}}
\caption{Accuracy of GPT-4o-mini, fine-tuned with 10,000 training instances, compared to our sampling and synthetic methods for improving the robustness of this closed-source model. $\uparrow$ and $\downarrow$ show better or worse average performance compared to the baseline, with $\dagger$ represents results where $p < 0.05$, using two-tailed bootstrapping statistical testing \cite{efron1993introduction}. The best results from fine-tuned models are in \textbf{bold}. See Appendix \ref{sec:appendix_std_dev} for standard deviations.}
\label{tab:main_sampling_results}
\end{table*}

\section{Results}

\subsection{Baselines} \label{sec:fewshot}
\paragraph{Naive fine-tuning reduces OOD performance.}
We find that the fine-tuned model $\mathcal{M}_\text{base}$ has substantially worse performance compared to the few-shot predictions from $\mathcal{M}$ on challenging OOD datasets. After fine-tuning, Challenge-OOD performance drops by 5.13\% for GPT-4o-mini and by 10.95\% for Gemini-2.0-Flash (Table \ref{tab:LLMs_with_and_without_FT}). The fine-tuning also results in large in-distribution improvements for both models. These results highlight how \textbf{fine-tuning can lead to large in-distribution improvements, but also substantially worse performance on challenging OOD datasets}. 
To the best of our knowledge, this is the first work to explicitly highlight this trade-off between in-distribution and OOD performance when fine-tuning LLMs instead of using in-context learning.
%
%In this work, we consider strategies to mitigate this loss of OOD performance while maintaining the in-distribution benefits from fine-tuning. 

\paragraph{Autoregressive LLMs are far stronger baselines than encoder models.}
Closed-source autoregressive LLMs substantially outperform smaller encoder models on OOD datasets (Table \ref{tab:Encoders_vs_LLMs}). This gap is particularly noticeable for the Challenge-OOD datasets, where autoregressive LLMs outperform encoder models by more than 10 percentage points. Notably, this robustness is achieved using only 10,000 training instances, which is less than 2\% of the 550k examples used to train each encoder model. Surprisingly, despite the large improvements in OOD performance, encoder models maintain strong performance on the in-distribution SNLI test set. \textbf{Our findings suggest that smaller encoder models are no longer appropriate baselines for measuring model robustness}, despite the continued use of these models in recent work \cite{korakakis-etal-2025-mitigating, DBLP:journals/ai/WangLWWSW25, cheng-amiri-2024-fairflow, honda-etal-2024-eliminate, koulakos-etal-2024-enhancing, stacey-rei-2024-distilling, zang2024explanation}. 

\subsection{Sampling Methods Lead to Improvements on Challenging OOD datasets} \label{sec:sampling_results}

Sampling more complex training examples leads to better performance on more challenging OOD datasets. %These examples can also yield small improvements on less challenging test sets, although not consistently.
Out of the methods tested,
\textit{Uncertainty Sampling} performs best, improving performance on Challenge-OOD by an average of 1.45\% (see Table \ref{tab:main_sampling_results}). Further experimentation shows improvements from Uncertainty Sampling when using either MNLI training data (Appendix \ref{appendix_subsec:mnli_experimentation}), or when the method is applied to challenging maths datasets (Appendix \ref{sec:appendix_maths_experiments}).
These improvements align with the intuition that instances with high prediction entropy represent challenging and under-represented cases in the training data. The inclusion of these examples helps models to generalise better to other challenging OOD instances. %Including these examples helps the model generalise beyond overly confident or shortcut-driven examples.

The other sampling methods result in smaller improvements. \textit{Concatenative Hypothesis Sampling} improves performance on Challenge-OOD by 1.15\%, while \textit{Difficulty Score Sampling} improves performance on Challenge-OOD by 0.84\%. 
On the other hand, \textit{Misclassified Sampling} does not improve performance on Challenge-OOD, with a qualitative analysis showing a large number of annotation errors when upsampling the misclassified training examples (see Section \ref{sec:analysis}).%suggesting why performance is worse than expected. 

We find that the improvements from these methods are only observed on more challenging OOD datasets. For Standard-OOD, the \textit{Uncertainty Sampling} and \textit{Difficulty Score Sampling} methods have similar performance to the baseline, contrasting with their improvements on Challenge-OOD. 
To further assess how our methods perform across test sets with varying levels of difficulty, we analyse the easy, ambiguous and hard MNLI-matched splits from \citet{cosma-etal-2024-hard}. Each sampling method has the largest improvement relative to the baseline on either the hard or the ambiguous splits (Appendix \ref{appendix_subsec:mnli_experimentation}). 
We conclude that \textbf{choosing more challenging training examples leads to better performance on challenging OOD datasets, but these improvements do not translate to less challenging OOD datasets}.

While the improvements from the sampling strategies are not significant when using 5 seeds (see Table \ref{tab:main_sampling_results}), we find that this is a result of the low sample size (and the high variance in OOD predictions from the sampling methods). Instead, when testing the best performing method, Uncertainty Sampling, with 20 seeds, we find statistically significant improvements on Challenge-OOD (see Appendix \ref{appendix_subsec:statistical_testing}).
To better understand the magnitude of our improvements, we also compare the improvements from our methods to prior work training on SNLI and evaluating on MNLI, a popular robustness setting (Table \ref{tab:snli_on_mnli}). Despite using a much better performing baseline, our improvements on MNLI are comparable in magnitude to previous work.

Finally, we explore variants of the best-performing methods. For \textit{Difficulty Score Sampling}, we test alternative scoring functions, and for \textit{Uncertainty Sampling}, we restrict upsampling to correctly predicted but uncertain examples. Neither adaptation further improves performance (see Appendix \ref{appendix_subsec:ablation_experiments}). 

\begin{table}
\centering
\resizebox{\columnwidth}{!}{
  \begin{tabular}{l @{\hskip 0.5em} c c c c c @{\hskip 0.5em}}
  \toprule
& & \multicolumn{2}{c}{MNLI-m} & \multicolumn{2}{c}{MNLI-mm} \\
Method & Baseline & Acc & Imp & Acc & Imp \\
  \midrule
  \rowcolor{light-gray} \multicolumn{6}{c}{Baselines (with OOD hyper-parameter tuning)}
  \\
% Negative sampling\textsuperscript{1} & LSTM & 43.76 & (2.10) & 43.66 & (3.91) \\
Hyp-only adversary\textsuperscript{1} & LSTM & 47.24 & \textbf{+1.38} & 49.24 & \textbf{+1.67} \\
Ensemble-adversaries\textsuperscript{2} & LSTM & 54.18 & +0.80 &  52.81 & (0.10) \\
Product of Experts\textsuperscript{3} & BERT  & 73.61 & (0.79) & 73.49 & (0.49) \\
Debiased Focal Loss\textsuperscript{3} & BERT & 73.58 & (0.82) & 74.00 & +0.02
\\

   \midrule
   \rowcolor{light-gray} \multicolumn{6}{c}{Baselines (no OOD hyper-parameter tuning)}\\
   ATA, EBD-Reg\textsuperscript{4} & BERT & 72.51 & +0.38 & 73.25 & +0.85 \\

Rationale supervision\textsuperscript{5} & BERT & 73.19 & +0.91 & 73.36 & +0.84 \\
NILE\textsuperscript{6} & RoBERTa & 77.07 & (2.22) & 77.22 & (2.07) \\
LIREx\textsuperscript{7} & RoBERTa & 79.85 & (0.27) & 79.79 & +0.06 \\
%KD$_{\text{ens}}$+aug\textsuperscript{ } & BERT & 76.42 & +1.41 & 76.45 & +1.48 \\
DTA$_{\text{ens}}$\textsuperscript{8} & DeBERTa & 85.77 & +1.21 & 86.18 & +1.40 \\
\midrule
   \rowcolor{light-gray} \multicolumn{6}{c}{Our best methods} \\
%Correct \& Uncertain & GPT-4o-mini & \textbf{88.21} & +1.28 & 88.24 & +1.00 \\
%Few-shot Question & GPT-4o-mini & 87.93 & +0.69 & 87.93 & +0.86 \\
%Hypothesis Mix & GPT-4o-mini & 86.55 & (0.38) & 86.69 & (0.55) \\
Uncertainty Sampling & GPT-4o-mini & \textbf{88.14} & +1.21 & 88.24 & +1.00 \\
%MNLI domain data & GPT-4o-mini & 87.79 & +0.86 & 88.34 & +1.10  \\
Short \& Simple Generation & GPT-4o-mini & 87.88 & +0.95 & \textbf{88.66} & +1.42 \\
%Gen data (4S, Simple) & GPT-4o-mini & 88.08 & +1.15 & \textbf{88.88} & +1.64 \\
Long \& Complex Generation & GPT-4o-mini & 87.74 & +0.81 & 88.07 & +0.83 \\
\bottomrule
\end{tabular}}
\caption{Our most successful methods compared to improvements from prior work. \textsuperscript{1}\citet{belinkov-etal-2019-dont}, \textsuperscript{2}\citet{stacey-etal-2020-avoiding}, 
\textsuperscript{3}\citet{karimi-mahabadi-etal-2020-end},
\textsuperscript{4}\citet{zang2024explanation},
\textsuperscript{5}\citet{Stacey_Belinkov_Rei_2022}, \textsuperscript{6}\citet{kumar-talukdar-2020-nile}, \textsuperscript{7}\citet{DBLP:conf/aaai/ZhaoV21}, \textsuperscript{8}\citet{stacey-rei-2024-distilling}.
}
\label{tab:snli_on_mnli}
\end{table}

\subsection{Data Generation Methods Improve Performance on Less Challenging OOD Data} \label{sec:data_generation_results}
\begin{table*}
\centering
\resizebox{\textwidth}{!}{
   \begin{tabular}{l c c c c c c c c @{\hskip 0.5em} c @{\hskip 0.5em} c @ {\hskip 0.5em} 
 c c c c}
  \toprule

& & \multicolumn{6}{c}{Challenge-OOD} & & \multicolumn{5}{c}{Standard-OOD}  \\ 
    \cmidrule{3-8} \cmidrule{10-14} 
    & SNLI & r1 & r2 & r3 \text{ } & COPA & INLI-I & WANLI & Avg. & MNLI-m & MNLI-mm & FEVER & Scitail & INLI-NLI & Avg. \\
  \midrule
  \rowcolor{light-gray} \multicolumn{15}{c}{GPT-4o-mini} \\
%Baseline w/o FT & 85.02 \text{   } \text{   } & 77.00 & 63.90 & 58.33 & 63.60 & 60.80 & 64.88 & 64.75 \text{   }  & 84.62 & 85.09 & 73.53 & 84.62 & 87.30 & 83.03 \text{   } \text{  } \\
%   \midrule
Baseline (10k FT) & \textbf{92.47} \text{   } \text{   } & 65.98 & 58.10 & 55.38 & 56.20 & \textbf{61.42} & 60.62 & 59.62 \text{ } \text{   } & 86.93 & 87.24 & 71.21 & 72.55 & 80.21 & 79.63 \text{  } \text{  } \\ % & 58.10 & 55.38 & \textbf{56.20} & \textbf{61.42} & \textbf{60.62} & 59.62 \text{ } \\%& 71.21 & 86.93 & 87.24 & 72.55 & 80.21 & 79.63 \text{ } \\
Short \& Simple Generation & \textbf{92.47} \text{   } \text{   } & 68.88 & 58.14 & 53.43 & \textbf{56.28} & 52.02 & \textbf{64.11} & 58.81$\downarrow$ \text{   } & \textbf{87.88} & \textbf{88.66} & \textbf{73.48} & \textbf{81.67} & 84.76 & \textbf{83.29}$\uparrow$$\dagger$ \\
Long \& Complex Generation & 92.26$\downarrow$ \text{    } & \textbf{69.34} & \textbf{60.28} & \textbf{57.42} & 55.90 & \textbf{61.42} & \textbf{60.86} & 60.87$\uparrow$$\dagger$ & 87.74 & 88.07 & 72.85 & 76.72 & \textbf{85.83} & 82.24 $\uparrow$$\dagger$\\
\midrule
  \rowcolor{light-gray} \multicolumn{15}{c}{Command R} \\
Baseline (10k) & 91.14 \text{  } \text{   } & \textbf{63.58} & \textbf{55.78} & 52.88 & \textbf{56.60} & \textbf{66.84} & 61.21 & \textbf{59.48} \text{  } \text{   } & 86.99 & 87.02 & 70.37 & 85.84 & 82.35 & 82.51 \text{   } \text{  }\\%& 55.78 & 52.88 & 56.60 & 66.84 & 61.21 & 59.48$\uparrow$ \\% & 58.04 & 86.99 & 87.02 & \textbf{85.84} & 82.35 & 80.05$\uparrow$ \\
Short \& Simple Generation & 91.04$\downarrow$ \text{   } & 61.48 & 52.50 & 50.23 & 54.30 & 58.10 & \textbf{62.11} & 56.45$\downarrow$ \text{   } & \textbf{87.79} & 87.68 & \textbf{71.09} & \textbf{88.47} & 83.19 & 83.64$\uparrow$ \text{   } \\

Long \& Complex Generation & \textbf{91.22}$\uparrow$ \text{   } & 63.20 & 55.04 & \textbf{53.02} & 53.54 & 63.24 & 61.40 & 58.24$\downarrow$ \text{   } & 87.62 & \textbf{87.75} & 71.06 & 86.66 & \textbf{85.79} & \textbf{83.78}$\uparrow$ \text{   } \\% & 54.34 & 52.12 & 55.00 & 64.60 & 60.02 & 58.12$\downarrow$ \\ %& \textbf{71.27} & 86.81 & \textbf{87.10} & 85.27 & \textbf{85.43} & \textbf{83.18}$\uparrow$ \\
%Improvement & +0.32 & (0.96) & (1.44) & (0.76) & (1.60) & (2.24) & (1.19) & (1.37) &  \\
 \midrule
   \rowcolor{light-gray} \multicolumn{15}{c}{Gemini-2.0-Flash} \\
%Baseline w/o FT & 82.39 \text{   }\text{   } & 79.70 & 71.50 & 67.08 & 92.30 & 84.60 & 65.42 & 76.85 \text{   } \text{   } & 83.24 & 82.45 & 72.90 & 72.58 & 84.60 & 79.15\text{   } \text{   } \\
%   \midrule
Baseline (10k) & \textbf{92.61} \text{   } \text{   } & 72.68 & 64.10 & 61.45 & 75.76 & 58.82 & 62.61 & 65.90 \text{ } \text{   }  & 88.30 & 88.54 & 71.56 & 73.42 & 85.09 & 81.38 \text{   } \text{   } \\ %& 64.10 & 61.45 & \textbf{75.76} & 58.82 & 62.61 & 65.90 \text{ } \\% & 71.56 & 88.30 & 88.54 & 73.42 & 85.09 & 81.38\text{ }  \\
Short \& Simple Generation  & 92.54$\downarrow$ \text{   } & 72.40 & 62.04 & 58.82 & 71.84 & 59.08 & 62.24 & 64.40$\downarrow$ \text{   } & 88.15 & 88.37 & 71.82 & \textbf{84.39} & 85.43 & \textbf{83.63}$\uparrow$$\dagger$ \\
Long \& Complex Generation & 92.59$\downarrow$ \text{   }  & \textbf{76.30} & \textbf{66.68} & \textbf{64.20} & \textbf{76.3}4 & \textbf{65.84} & \textbf{62.96} & \textbf{68.72}$\uparrow$$\dagger$ & \textbf{88.95} & \textbf{89.23} & \textbf{72.95} & 78.95 & \textbf{87.59} & 83.53$\uparrow$$\dagger$ \\
\bottomrule
  \end{tabular}}
\caption{Testing our data generation methods for GPT-4o-mini, Command R and Gemini-2.0-Flash. As Command R and Gemini do not provide probability scores, we do not also test our Uncertainty Sampling method. $\dagger$ represents results where $p < 0.05$, using two-tailed bootstrapping statistical testing \cite{efron1993introduction}. The best results from fine-tuned models are in \textbf{bold}.
}
\label{tab:unlab_data_table}
\end{table*}

Replacing some of the human annotated training examples with LLM-generated synthetic data substantially improves performance on Standard-OOD, with gains ranging from 2.61\% to 3.66\% across each method. Table \ref{tab:main_sampling_results} shows that every Standard-OOD test set benefits from the inclusion of the synthetic data, and that every method results in statistically significant improvements.
We also test our data generation methods with Command R\footnote{Due to slower few-shot inference with Command R, we fine-tune this model with the data generated by GPT-4o-mini} and Gemini-2.0-Flash (Table \ref{tab:unlab_data_table}), finding further performance improvements on Standard-OOD. 
These improvements mostly do not translate to the Challenge-OOD datasets, where the inclusion of the synthetic data mostly results in worse performance compared to the baseline (see Short \& Simple Generation, Table \ref{tab:unlab_data_table}). This is due to the relatively simple entailment relationships provided in the synthetic data, which we analyse in Section \ref{sec:analysis}. 

Prompting strategies can increase the complexity of the synthetic data, and our Long \& Complex Generation successfully improves performance on Challenge-OOD for GPT-4o-mini and Gemini-2.0-Flash (see Table \ref{tab:unlab_data_table}), with improvements of up to 2.82\% compared to the baseline. We conclude that \textbf{the inclusion of LLM-generated synthetic data leads to large OOD improvements, with further gains observed on more challenging OOD datasets when increasing the data complexity}.

We also test whether the \textit{if in doubt, discard} validation method is necessary for strong performance when using LLM-generated data. We find that the additional validation is most beneficial for more challenging synthetic examples, improving performance on  Challenge-OOD for the Long \& Complex Generation method (+2.26\%; see Appendix \ref{appendix_subsec:ablation_experiments}).
Finally, when training on MNLI instead of SNLI, we find that improvements are limited to Challenge-OOD (see Long \& Complex Generation, Appendix \ref{appendix_subsec:mnli_experimentation}), suggesting that the additional synthetic data is most helpful for single domain datasets.

\subsection{Qualitative analysis}\label{sec:analysis}
\paragraph{Sampling more difficult examples results in more label noise.}
We analyse the examples selected for $\mathcal{D}_\text{up}$ by manually inspecting 50 examples per method and assigning a difficulty score between 1-10, and reviewing the label annotations for each method.\footnote{We describe the differences between the labels and the judgements from one of the paper authors as `annotation errors', however labelling in NLI can be highly subjective}
\textit{Misclassified Sampling} examples have the highest difficulty score (5.92), but we find this also corresponds to the most annotation errors (54\%). While Uncertainty Sampling produces less complex instances (5.26), there are less than half the number of annotation errors (24\%). Despite this difference, we see a large overlap between the two methods, with 33.1\% of $\mathcal{D}_\text{up}$ shared between both methods. We suggest that Uncertainty Sampling retains many of the valuable examples from Misclassified Sampling while avoiding its high rate of label errors.
Difficulty Score Sampling has a lower average difficulty score (4.44), but with even fewer annotation errors (16\%). In comparison, the difficulty score of the original SNLI training data was 3.84, with only 4\% of instances with annotation errors.

\paragraph{Generation methods can result in simpler instances that are more likely to be correctly labelled.}
The Short \& Simple Generation produces less difficult examples than the original training data (3.40 vs 3.84), with 4\% of annotation errors. This low complexity helps to explain why training with this synthetic data does not improve Challenge-OOD performance.
We find that prompting strategies can successfully increase the complexity of the synthetic data. In particular, the Long \& Complex Generation method produces more challenging examples (with a difficulty score of 4.52). As expected, there are also a larger number of annotation errors with the more challenging synthetic data (12\%).

\paragraph{Domain-specific synthetic examples do not yield domain-specific improvements}
When generating synthetic data, we choose domains not present in the OOD test sets that we evaluate on. Surprisingly, generating data for the domains contained in an OOD test set does not further improve performance on that specific test set. For example, both the MNLI Domains Generation (which uses the MNLI-matched domains) and the Short \& Simple Generation (which does not) achieve similar accuracy on Challenge-OOD and Standard-OOD, with nearly identical performance on MNLI-matched. This is despite the MNLI Domains Generation being generated specifically for these same domains. 
Furthermore, the MNLI Domains Generation results in larger improvements compared to the baseline on MNLI-mismatched (containing unseen domains) than on MNLI-matched (which contains the same domains).

\section{Conclusion}

We examine the effect of fine-tuning closed-source LLMs on OOD performance for NLI and find that the substantial performance improvements from fine-tuning are often offset by large drops in OOD performance, particularly on challenging OOD NLI datasets. 
Due to the lack of mitigation methods compatible with closed-source models, we introduce data selection strategies that are compatible with the API training constraints of these models. Our data selection methods improve OOD NLI performance while keeping within a fixed training budget.

We find that prioritising more complex examples for fine-tuning improves performance on challenging OOD datasets, while replacing a portion of the training set with LLM-generated examples improves performance on less challenging datasets. 
To overcome the lack of complexity in the LLM-generated synthetic data, we introduce prompting strategies to make this data more complicated. Training with more challenging synthetic data results in improvements across almost all of the OOD test sets analysed, including both challenging and less challenging datasets. This highlights the promise of training with synthetic data when it is both diverse and sufficiently challenging.
%
% These LLM-generated instances offer diversity across different domains, but without any additional complexity.

Finally, we find that recent autoregressive LLMs are substantially more robust than encoder models, despite the similar in-distribution performance between these models. We therefore suggest that autoregressive LLMs are a more appropriate baseline for future work on model robustness. %despite having similar performance in-distribution.

%Overall, we find there is no one-size-fits-all solution for improving OOD performance, with the best approach depending on the complexity of the OOD data.

\section*{Limitations}

Despite the manageable cost for fine-tuning a single closed-source model with 10,000 training instances (approximately \$5 for GPT-4o-mini), replicating the full experimentation in this work would be much more costly. It also is possible that the LLMs may have had access to some of the out-of-distribution datasets during training, however this is a widespread issue across the field, with many open-weights models also not disclosing their full training data. 

We suggest in Section \ref{sec:fewshot} that smaller encoder models such as RoBERTa or DeBERTa are no longer appropriate baseline models, however we also recognise that there may still be circumstances when encoder models are still helpful, for example if only considering in-distribution performance. %
While the model size is one factor that explains the better performance of recent autoregressive LLMs compared to encoder models, current LLMs also undergo much more pre-training than older encoder models, and are more developed after receiving considerably more research attention over recent years. 
It is possible that future work may improve the robustness of encoder models, although the recently proposed ModernBERT \cite{warner2024smarterbetterfasterlonger} is still outperformed by DeBERTa-v3-large on GLUE \cite{wang-etal-2018-glue}.

We choose to focus our experimentation on NLI, providing a comprehensive out-of-distribution evaluation of task performance. This is motivated by the large body of existing work studying model robustness on this task. While our findings are specific to improving the robustness of closed-source NLI models, they may also generalise to other tasks. This is supported by our experiments using the Uncertainty Sampling method on mathematics datasets, which shows promising results. The best robustness strategy may also depend on the task, with other tasks (such as mathematics tasks) potentially not having the same challenges with annotation errors or subjective labelling as is the case with NLI.

Finally, we use either SNLI or MNLI as our training data, testing performance on datasets with a similar level of difficulty (Standard-OOD) or on more challenging datasets (Challenge-OOD). We choose SNLI and MNLI as we aim to investigate improving performance for the common real-world setting where the training data does not represent the full breadth of possible domains or data complexity, and therefore leads to decreased OOD performance. In other settings where the training data already contains sufficiently diverse or complex data, this problem is greatly reduced, and data selection methods would likely not have the same effect. %Our results are therefore true with respect to our SNLI or MNLI training data.

\section*{Acknowledgements}

We would like to thank Nikolai Rozanov for his helpful ideas about this work. Joe Stacey was supported by the Apple Scholars in AI/ML PhD fellowship.

\bibliography{custom}

\appendix

\section{Related 
Work}\label{sec:extra_related_work}
\paragraph{Model debiasing.}
Model centric debiasing methods for NLI mostly involve either weighting the loss of training examples based on a bias model's predictions \cite{karimi-mahabadi-etal-2020-end, clark-etal-2019-dont, ghaddar-etal-2021-end, DBLP:conf/ijcai/DuYCLZSWW023, DBLP:journals/corr/abs-2211-13331, DBLP:conf/nips/XiongCPCML21}, adding the bias model log probabilities, logits, or attention scores during training \cite{karimi-mahabadi-etal-2020-end, clark-etal-2019-dont, he-etal-2019-unlearn, DBLP:conf/iclr/Sanh0BR21, wang-etal-2023-robust, DBLP:conf/nips/XiongCPCML21}, adjusting the model's soft probabilities at inference time based on the predictions of a biased model \cite{udomcharoenchaikit-etal-2022-mitigating, Tian_Cao_Zhang_Xing_2022}, or encouraging a model to be less confident when a bias model or a perturbed input can result in correct predictions \cite{utama-etal-2020-mind, cheng-amiri-2024-fairflow, du-etal-2021-towards}. Each of these methods accommodate a variety of ways of choosing the bias model, which include identifying specific dataset biases \cite{he-etal-2019-unlearn, clark-etal-2019-dont, DBLP:conf/ijcai/DuYCLZSWW023} including the hypothesis-only NLI bias \cite{karimi-mahabadi-etal-2020-end, he-etal-2019-unlearn}, using models that are not large enough \cite{DBLP:conf/iclr/Sanh0BR21, clark-etal-2020-learning} or do not contain enough training data \cite{utama-etal-2020-towards}, or finally, using features from earlier layers in the model \cite{DBLP:journals/ai/WangLWWSW25, ghaddar-etal-2021-end}. 

For LSTM models \cite{hochreiter1997long}, adversarial training has also been effective at removing dataset biases, either removing these biases from a model representations \cite{belinkov-etal-2019-dont, belinkov-etal-2019-adversarial, stacey-etal-2020-avoiding} or embeddings \cite{zhou-bansal-2020-towards}. Despite the success of adversarial training with LSTM models, we are not aware of any work showing that these methods can also be successfully applied to transformer-based NLI models. 

\paragraph{Data augmentation.}
An increasingly common approach for improving model robustness involves exploiting recent advances in LLMs to generate synthetic datasets, creating large volumes of additional training data. This additional training data has proved helpful, both for improving robustness \cite{hosseini-etal-2024-synthetic, chen-etal-2023-disco, wang-etal-2023-ibadr, liu-etal-2022-wanli, wu-etal-2022-generating} and improving in-distribution performance \cite{DBLP:journals/corr/abs-2412-09263}. Instead of using LLMs to generate additional training instances, new examples can also be created by perturbing existing training examples \cite{minervini-riedel-2018-adversarially, zhou-bansal-2020-towards, liu-etal-2020-empirical, min-etal-2020-syntactic}. This can involve paraphrasing \cite{zhou-bansal-2020-towards}, swapping the subject and object in the input sentences \cite{min-etal-2020-syntactic}, adding additional text at the end of the inputs \cite{zhou-bansal-2020-towards}, swapping the hypothesis and premise \cite{liu-etal-2020-empirical}, or using first-order logic rules \cite{minervini-riedel-2018-adversarially}. New examples can also be created using human annotation \cite{DBLP:conf/iclr/KaushikHL20, yanaka-etal-2019-help, he-etal-2023-targeted}. 

Whether human-annotated or LLM-generated,  augmented data used to improve robustness is either: 1) created by making small changes to the hypothesis or premise \cite{chen-etal-2023-disco, yanaka-etal-2019-help, min-etal-2020-syntactic, wen-etal-2022-autocad}, 2) created to be similar to challenging or unbiased examples in the training domain \cite{liu-etal-2022-wanli, wu-etal-2022-generating, he-etal-2023-targeted, wang-etal-2023-ibadr},
or 3) created as entirely new instances for different domains \cite{hosseini-etal-2024-synthetic, stacey-rei-2024-distilling}.

\paragraph{Data filtering.}
Rather than augment a dataset with additional examples, a dataset can be filtered to remove biased examples \cite{wu-etal-2022-generating, DBLP:conf/icml/BrasSBZPSC20}, helping to improve robustness. In its simplest form, dataset filtering can involve a filtering process to remove examples that have been identified as having a sufficient level of bias \cite{ravichander-etal-2023-bias}. \citet{wu-etal-2022-generating} introduce an algorithm for iteratively removing biased instances, considering whether hand-crafted features correlate to a specific class using z-statistics \cite{gardner-etal-2021-competency}. Alternatively, \citet{DBLP:conf/icml/BrasSBZPSC20} iteratively partition a dataset into a train and test partition, training on the training examples, before evaluating on the test examples. This process is repeated, each time choosing different train and test splits, with examples filtered out which are most commonly predicted as the correct label when they are in the test split. This process is repeated for a set number of steps, resulting in a reduced, less biased dataset. 
Beyond NLI, filtering out lower quality training examples has been found to increase both model performance \cite{ijcai2025p0928, pang2025token, wang-etal-2025-data-whisperer} and robustness \cite{hu2025donodefficientgeneralizableinstruction}.

\paragraph{Improving robustness with explanations.}

Recent work has also investigated whether NLI robustness can be improved with the use of explanations, using human annotated explanations during training \cite{Stacey_Belinkov_Rei_2022, ross-etal-2022-self, koulakos-etal-2024-enhancing, zang2024explanation} with the e-SNLI human annotated explanations \cite{DBLP:conf/nips/CamburuRLB18}. Models can learn from the human explanations by supervising the CLS token in the final layer of the model \cite{Stacey_Belinkov_Rei_2022}, or by training models to generate explanations alongside making class label predictions \cite{ross-etal-2022-self, koulakos-etal-2024-enhancing, ludan-etal-2023-explanation}. However, not all past work using explanations during training has led to improvements in robustness \citep{kumar-talukdar-2020-nile, DBLP:conf/aaai/ZhaoV21, DBLP:conf/nips/CamburuRLB18}. Recent work has also tested whether including chain of thought explanations with in-context examples improves robustness \cite{mueller-etal-2024-context, he-etal-2024-using}, with mixed results. 

\paragraph{Learning from minority examples.}
Minority examples can be defined as instances that counter common spurious patterns in a dataset \cite{tu-etal-2020-empirical}. Modifying the training process can help models to learn more from these minority examples, improving out-of-distribution performance for NLI \cite{korakakis-vlachos-2023-improving, DBLP:conf/icml/LiuHCRKSLF21, yaghoobzadeh-etal-2021-increasing}. \citet{DBLP:conf/icml/LiuHCRKSLF21} test a fine-tuned model on its own training data, identifying minority examples as the instances misclassified in the training set. These minority examples were upsampled during a second round of training. Similarly, \citet{yaghoobzadeh-etal-2021-increasing} identify training examples that were misclassified, but also include examples that were predicted as the wrong class at some point during training. A second phase of training then trained only on these minority examples. Rather than identifying minority examples as those that a single model has misclassified, ensembles can also be used to better identify these examples, with minority examples defined as any training instance that any model in an ensemble has incorrectly predicted \cite{stacey-rei-2024-distilling}.

\paragraph{Improving robustness with knowledge distillation.}

NLI robustness has also been studied in the context of knowledge distillation, where the distillation process is modified to produce more robust student models. This can involve smoothing the teacher soft predictions \cite{du-etal-2023-robustness, jafari-etal-2021-annealing}, or applying methods inspired by Just Train Twice \cite{DBLP:conf/icml/LiuHCRKSLF21} to learn more from challenging examples \cite{du-etal-2023-robustness, stacey-rei-2024-distilling}. Using additional unlabelled examples can also improve the robustness of NLI student models, generating additional examples by masking individual words and adversarially generating words to replace these masks \cite{rashid-etal-2021-mate,li-etal-2021-select-one, haidar-etal-2022-cilda}, or using LLMs to generate unlabelled data that can be used during distillation \cite{stacey-rei-2024-distilling}.

\begin{table*}[ht]
\centering
\small
\begin{tabular}{lp{10cm}}
\toprule
\textbf{Symbol} & \textbf{Description} \\
\midrule
$\mathcal{D}$ & Full NLI dataset (all available annotated examples). \\
$\mathcal{D}_{\text{init}}$ & Initial training subset used to fine-tune the base model ($|\mathcal{D}_{\text{init}}| = m$). \\
$\mathcal{D}_{\text{potential}}$ & Candidate examples from $\mathcal{D} \setminus \mathcal{D}_{\text{init}}$ considered for upsampling. \\
$\mathcal{D}_{\text{up}}$ & New, challenging examples selected to improve robustness. \\
$\mathcal{D}_{\text{down}}$ & Subset of $\mathcal{D}_{\text{init}}$ removed to make room for $\mathcal{D}_{\text{up}}$, preserving fixed budget. \\
\bottomrule
\end{tabular}
\caption{Notation summary for dataset subsets used in our training strategies.}
\label{tab:dataset_notation}
\end{table*}

\paragraph{Distributionally robust optimization.}
\citet{DBLP:conf/iclr/SagawaKHL20} demonstrate how Distributionally Robust Optimization (DRO) can improve the robustness of NLI models, training models with the objective of minimising the worst-group loss. \citet{DBLP:conf/iclr/SagawaKHL20} segment MNLI instances into groups based on the class (entailment, neutral or contradiction), and whether there is a negation term present, creating six groups from these different possible combinations. During training, the loss for instances in each group are adjusted, with this adjustment depending on the loss so far during training for the groups they belong to. \citet{DBLP:conf/iclr/SagawaKHL20} find that DRO improves the robustness on instances belonging to the worst group during inference, although with lower overall performance. Rather than using known heuristics, subsequent work has identified automatic methods for identifying the training groups \cite{DBLP:journals/corr/abs-2204-13749, DBLP:conf/icml/BaoCB21, DBLP:conf/iclr/ParanjapeDSZH23}. The worst-group performance tested by \citet{DBLP:conf/iclr/SagawaKHL20} can also be improved by simply adjusting the class imbalance between groups in the training data \cite{DBLP:journals/corr/abs-2110-14503}.

% AGRO: ADVERSARIAL DISCOVERY OF ERROR-PRONE GROUPS FOR ROBUST OPTIMIZATION (to include above)

\paragraph{Atomic decomposition methods.}

The task of NLI can be decomposed into different atoms, with predictions made separately for individual atoms. These predictions are then aggregated into instance-level model predictions, with some evidence this can lead to robustness improvements. The atom-level decomposition can involve segmenting hypotheses into spans, and making predictions for each hypothesis span independently, which can improve robustness when there is limited training data  \cite{stacey-etal-2022-logical}. Alternatively, decomposing each premise into a list of atomic facts, and making predictions for each premise fact independently can result in out-of-distribution improvements \cite{stacey-etal-2024-atomic}. 

Rather than decompose each hypothesis or premise, \citet{DBLP:journals/corr/abs-2502-09567} train a model to edit (or \textit{morph}) a premise step by step until it matches the hypothesis, using a fine-tuned NLI model to track how entailment changes over these intermediate steps. During inference, model predictions are aggregated based on the entailment predictions at each step, decomposing the task into a series of smaller inference steps. This decomposition process was found to improve the model's generalisation abilities. Symbolic methods can also improve robustness, with \citet{kalouli-etal-2020-hy} using a symbolic approach (combined with a standard neural model) to make predictions based on the differences between the premise and hypothesis. \citet{opitz-etal-2023-amr4nli} also find robustness improvements using a hybrid approach, combining BERT model predictions with a score derived from Abstract Meaning Representation (AMR) graphs \cite{banarescu-etal-2013-abstract} of the hypothesis and premise.

\paragraph{Further NLI robustness strategies.}

Further strategies to improve NLI robustness include: supervising gradients of instances based on the difference between each instance and a corresponding counterfactual \cite{DBLP:conf/eccv/TeneyAH20}, using Variational Information Bottleneck's to improve the robustness of models in a low resource setting \cite{DBLP:conf/iclr/MahabadiBH21}, and introducing a mixture of experts model that trains different classifiers on top of a BERT encoder, weighting the soft probabilities of each expert \cite{honda-etal-2024-eliminate}. Finally, \citet{DBLP:journals/eswa/LeeKPL25} find that using prompt tuning can improve robustness compared to other fine-tuning strategies, tuning the prompt tokens that have the least importance for dev set predictions.

\section{Notation}
\label{sec:notation}

To aid readability, we provide a summary of the dataset subsets in Table~\ref{tab:dataset_notation} used throughout our methods section. This includes the roles of each subset in training and evaluation.

\begin{table*}
\centering
\resizebox{\textwidth}{!}{
 \begin{tabular}{l c c c c c c c c @{\hskip 0.5em} c @{\hskip 0.5em} c @ {\hskip 0.5em} 
 c c c c}
  \toprule
& & \multicolumn{6}{c}{Challenge-OOD} & & \multicolumn{5}{c}{Standard-OOD}  \\ 
    \cmidrule{3-8} \cmidrule{10-14} 
    & SNLI & r1 & r2 & r3 \text{ } & COPA & INLI-I & WANLI & Avg. & MNLI-m & MNLI-mm & FEVER & Scitail & INLI-NLI & Avg. \\
  \midrule
  \rowcolor{light-gray} \multicolumn{15}{c}{GPT-4o-mini baselines} \\
5k baseline & 92.18 & 65.92 & 57.70 & 55.68 & 56.40 & 61.04 & 61.51 & 59.71 & 86.93 & \textbf{87.46} & 71.45 & 76.54 & 82.38 & 80.95 \\% & 57.70 & 55.68 & 56.40 & 61.04 & 61.51 & 59.71 \\% & 71.45 & 86.93 & 87.46 & 76.54 & 82.38 & 80.95 \\
10k baseline & 92.47 & 65.98 & 58.10 & 55.38 & 56.20 & \textbf{61.42} & 60.62 & 59.62 & 86.93 & 87.24 & 71.21 & 72.55 & 80.21 & 79.63 \\% & 58.10 & 55.38 & 56.20 & 61.42 & 60.62 & 59.62 \\% & 71.21 & 86.93 & 87.24 & 72.55 & 80.21 & 79.63 \\
20k baseline & \textbf{92.78} & 66.06  & 57.54 & 54.83 & 58.98 & 60.54 & 59.46 & 59.57 & \textbf{87.32} & 87.10  & 71.20 & 69.65 & 80.61 & 79.18 \\% & 57.54 & 54.83 & 58.98 & 60.54 & 59.46 & 59.57 \\% & 71.20 & 87.32 & 87.10 & 69.65 & 80.61 & 79.18 \\
Few-shot baseline & 85.02 & \textbf{77.00} & \textbf{63.90} & \textbf{58.33} & \textbf{63.60} & 60.80 & \textbf{64.88} & \textbf{64.75} & 84.62 & 85.09 & \textbf{73.53} & \textbf{84.62} & \textbf{87.30} & \textbf{83.03} \\ %& 63.90 & 58.33 & 63.60 & 60.80 & 64.88 & 64.75 \\% & 73.53 & 84.62 & 85.09 & 84.62 & 87.30 & 83.03 \\
\bottomrule
  \end{tabular}}
\caption{We compare our baseline with 10k training instances to a baseline with twice this amount, showing that more training examples has led to better robustness, but worse performance on SNLI-test. As expected, we also show that our baseline has substantially better performance than a few-shot baseline $\mathcal{M}$. Unlike all previous experiments, the few-shot baseline uses a single seed rather than an average of 5 seeds.}
\label{tab:appendix_baseline_table}
\end{table*}
\begin{table*}
\centering
\resizebox{\textwidth}{!}{
 \begin{tabular}{l c c c c c c c c @{\hskip 0.5em} c @{\hskip 0.5em} c @ {\hskip 0.5em} 
 c c c c}
  \toprule

& & \multicolumn{6}{c}{Challenge-OOD} & & \multicolumn{5}{c}{Standard-OOD}  \\ 
    \cmidrule{3-8} \cmidrule{10-14} 
    & SNLI & r1 & r2 & r3 \text{ } & COPA & INLI-I & WANLI & Avg. & MNLI-m & MNLI-mm & FEVER & Scitail & INLI-NLI & Avg. \\
  \midrule
  \rowcolor{light-gray} \multicolumn{15}{c}{GPT-4o-mini} \\

Without fine-tuning & 85.02 \text{  } & \textbf{77.00} & \textbf{63.90} & \textbf{58.33} & \textbf{63.60} & 60.80 & \textbf{64.88} & \textbf{64.75} \text{  } & 84.62 & 85.09 & \textbf{73.53} & \textbf{84.62} & \textbf{87.30} & \textbf{83.03} \text{  } \\
With fine-tuning (10k) & \textbf{92.47}$\uparrow$ & 65.98 & 58.10 & 55.38 & 56.20 & \textbf{61.42} & 60.62 & 59.62$\downarrow$ & 86.93 & 87.24 & 71.21 & 72.55 & 80.21 & 79.63$\downarrow$  \\
  \midrule
    \rowcolor{light-gray} \multicolumn{15}{c}{Gemini-2.0-Flash} \\

Without fine-tuning & 82.39 \text{ } & \textbf{79.70} & \textbf{71.50} & \textbf{67.08} & \textbf{92.30} & \textbf{85.10} & \textbf{65.42} & \textbf{76.85} \text{ } & 83.24 & 82.45 & \textbf{72.90} & 72.58 & 84.60 & 79.15 \text{ } \\
With fine-tuning (10k) & \textbf{92.61}$\uparrow$ & 72.68 & 64.10 & 61.45 & 75.76 & 58.82 & 62.61 & 65.90$\downarrow$  & \textbf{88.30} & \textbf{88.54} & 71.56 & \textbf{73.42} & \textbf{85.09} & \textbf{81.38}$\uparrow$ \\
\bottomrule
  \end{tabular}}
\caption{Accuracy of both GPT-4o-mini and Gemini-2.0-Flash both before and after fine-tuning with 10,000 training instances. Fine-tuning substantially improves in-distribution performance on SNLI, but at the expense of OOD performance on challenging NLI datasets}
\label{tab:appendix_with_and_without_FT}
\end{table*}

\section{Experimental Details} \label{appendix_sec:experimental_details}

%\subsubsection{Hyper-parameter Tuning} 
\paragraph{Hyper-parameter Tuning.}
\label{appendix_subsec:hyper_param_tuning}
When fine-tuning our encoder models, for DeBERTa-base and DeBERTa-large we use a learning rate of $10^{-6}$, training for two epochs, while for RoBERTa-large we use a learning rate of $5 \times 10^{-6}$. %following \citet{stacey-rei-2024-distilling}. 
For GPT-4o-mini, Command R and Gemini-2.0-Flash, we use the default fine-tuning parameters. We do this to consider a realistic scenario where additional funds may not be available for further hyper-parameter tuning. For GPT-4o-mini, this involves training for 2 epochs, with the learning rate multiplier set to 1.8. For Command R, we fine-tune for a single epoch (with a learning rate of 0.01), while for Gemini-2.0-Flash we also use the default of 13 epochs.

We use the following checkpoints for each model:  gpt-4o-mini-2024-07-18 for GPT-4o-mini, gemini-2.0-flash-001 for Gemini, and the `Base Type Chat' model for fine-tuning Command R. %For Command R, we consider command-r-plus-08-2024 to be an equivalent model for few-shot predictions before the fine-tuning.

When using $\mathcal{M}$ to make few-shot predictions, we use three in-context samples, each with chain of thought reasoning.

\paragraph{Creating COPA-NLI.} \label{sec:appendix_copa_nli}
Instances in the Balanced-COPA dataset consist of a premise, and two choices, with the task of choosing which of the two choices is the most like cause or effect of the premise. For COPA-NLI, we use the same premise, but incorporating both choices into the hypothesis statement. When the task is to determine the cause of the premise, we use the following template for the hypothesis: ``[choice\_1]'' is a more likely cause of this than ``[choice\_2]'', with a label of entailment or non-entailment. Then, for instances when we determine which sentence is the most likely effect of the premise, we use the following hypothesis template: ``[choice\_1]'' is a more likely effect of this than ``[choice\_2]''. We duplicate each instance in Balanced-COPA, changing the order of the two choices in the hypothesis and changing the NLI label.

\section{Further Experiments}\label{appendix_sec:further_experiments}

\subsection{Baseline Analysis}
\label{appendix_subsec:baseline_analysis}
We experiment with training our baseline model $\mathcal{M}_\text{base}$ with 5k, 10k and 20k training instances, to understand whether training with more data improves in-distribution or out-of-distribution performance (Table \ref{tab:appendix_baseline_table}). Overall, we find that more fine-tuning data improves in-distribution performance, but without improvements in model robustness. Our few-shot baseline using $\mathcal{M}$ is substantially worse on SNLI-test, but also substantially more robust on Standard-OOD and Challenge-OOD.

We also include the full results when comparing GPT-4o-mini and Gemini-2.0-Flash with and without fine-tuning (see Table \ref{tab:appendix_with_and_without_FT}).

\begin{table*}
\centering
\resizebox{\textwidth}{!}{
  \begin{tabular}{l @{\hskip 0.5em} c c c c c@{\hskip 0.5em} c c c c c@{\hskip 0.3em} c c c c}
  \toprule

& & \multicolumn{6}{c}{Challenge-OOD} & & \multicolumn{5}{c}{Standard-OOD}  \\ 
    \cmidrule{3-8} \cmidrule{10-14} 
    & MNLI-m & r1 & r2 & r3 & COPA & INLI-I & WANLI & Avg. & SNLI-dev & SNLI-test & FEVER & Scitail & INLI-NLI & Avg. \\
  \midrule
  \rowcolor{light-gray} \multicolumn{15}{c}{GPT-4o-mini - MNLI training data} \\
Baseline (10k) & 90.73 & 68.72 & 56.56 & 52.58 & 50.08 & 47.10 & 65.54 & 56.76\text{ } & \textbf{91.81} & \textbf{91.38} & 73.18 & 85.58 & 86.05 & 85.60 \text{ } \\
Uncertainty Sampling & \textbf{91.09} & \textbf{70.80} & \textbf{58.68} & \textbf{54.53} & 50.02 & 49.48 & 65.57 & 58.18$\uparrow$ & 91.43 & 91.10 & \textbf{73.45} & 85.52 & 86.53 & 85.61$\uparrow$ \\
Short \& Simple Generation & 90.45 & 68.17 & 55.84 & 52.08 & 50.14 & 46.28 & \textbf{66.45} & 56.50$\downarrow$ & 91.38 & 91.08 & 73.42 & 85.79 & 86.08 & 85.55$\downarrow$ \\
Long \& Complex Generation & 90.60 & 69.06 & 57.32 & 54.48 & \textbf{50.24} & \textbf{53.26} & 65.54 & \textbf{58.32}$\uparrow$ & 91.68 & 91.23 & 73.22 & \textbf{85.81} & \textbf{87.33} & \textbf{85.86}$\uparrow$ \\

\bottomrule
  \end{tabular}}
\caption{We experiment with our best performing methods when training on MNLI, a dataset with five different training domains.}
\label{tab:mnli_results_table}
\end{table*}
\subsection{MNLI Experimentation} \label{appendix_subsec:mnli_experimentation}

\paragraph{Training with MNLI.}
We additionally experiment with using MNLI as our training data instead of SNLI, while replacing the MNLI-matched and MNLI-mismatched validation sets in Standard-OOD with SNLI-dev and SNLI-test. Similar to our results with the SNLI training data, we find improvements on Challenge-OOD when using Uncertainty Sampling (+1.42, see Table \ref{tab:mnli_results_table}). These improvements are of a similar magnitude to the improvements observed when training on SNLI (+1.45, see Table \ref{tab:main_sampling_results}). Using the Uncertainty Sampling method with MNLI training data also maintains the baseline performance on Standard-OOD (+0.01\%, see Table \ref{tab:mnli_results_table}).

We find that the Long \& Complex Generation method also improves performance on Challenge-OOD (+1.56, see Table \ref{tab:mnli_results_table}). This is a larger improvement compared to our results training on SNLI (+1.25, see Table  \ref{tab:main_sampling_results}). However, unlike our experiments training on SNLI, we do not see the same performance improvements on Standard-OOD from our data generation methods. As the MNLI training data already contains data from five different domains, we suggest that including additional, diverse LLM-generated training data is less beneficial in this case.

\paragraph{Evaluation on MNLI difficulty splits.}

We use the hard, ambiguous and easy splits provided by \citet{cosma-etal-2024-hard} to understand why our methods have a performance advantage when evaluated on MNLI, and whether this is driven by improvements on more challenging examples. Table \ref{tab:snli_hard_mnl_hard} shows how the out-of-distribution improvements on MNLI from our sampling methods are largest on either the hard or the ambiguous test splits. For Misclassified Sampling and Difficulty Score Sampling, the largest performance improvement is on the hard test set, whereas for Concatenative Hypothesis Sampling and Uncertainty Sampling the largest increase is on the ambiguous test set. For the data generation methods, we consistently see improvements on the easy test split. However, surprisingly, there are also improvements on the hard test splits when using the LLM-generated synthetic data. %This may be because MNLI is a single sentence dataset, so the complexity we introduce in the Long \& Complex Generation may not be relevant for this dataset.

As a comparison, we also provide the results for our methods on the difficulty splits for SNLI. In this case, the differences between the baseline and our methods are smaller, with our methods providing no clear advantage over the baseline model.

\begin{table}
\centering
\resizebox{\columnwidth}{!}{
  \begin{tabular}{l @{\hskip 0.5em} c c c c c c @{\hskip 0.5em}}
  \toprule
& \multicolumn{3}{c}{SNLI} & \multicolumn{3}{c}{MNLI-m} \\
Method & Hard & Amb. & Easy & Hard & Amb. & Easy \\
  \midrule
   \rowcolor{light-gray} \multicolumn{7}{c}{Baselines}\\
GPT-4o-mini w/o FT \\
GPT-4o-mini (10k FT) & 67.98 & 89.68 & 97.47 & 59.27 & 80.90 & 91.89 \\
\midrule
  \rowcolor{light-gray}  \multicolumn{7}{c}{Our methods} \\
%Correct \& Uncertain & 67.32 & 90.33 & 97.72 & 61.56 & \textbf{82.45} & 92.98 \\
Misclassified Sampling & \textbf{68.20} & 89.44 & 97.27 & 60.49 & 80.65 & 91.69 \\
Difficulty Score Sampling & 67.85 & 89.98 & 97.68 & 61.30 & 81.25 & 92.49 \\
Concat Hypothesis Sampling & 67.80 & \textbf{90.18} & 97.50 & 58.30 & \textbf{81.56} & 91.39 \\
Uncertainty Sampling & \textbf{68.20} & 90.04 & \textbf{97.81} & 61.01 & 82.31 & 92.99 \\
Short and Simple Generation & 68.15 & 89.70 & 97.43 & \textbf{64.42} & 77.55 & \textbf{93.08} \\
Long \& Simple Generation & 67.97 & 89.69 & 97.56 & 63.65 & 79.54 & \textbf{93.08} \\
Long \& Complex Generation & 68.41 & 89.14 & 97.23 & 60.06 & 80.73 & 92.89 \\
\bottomrule
  \end{tabular}}
\caption{Performance of our methods on Hard, Ambiguous and Easy test splits of SNLI and MNLI-matched.}
\label{tab:snli_hard_mnl_hard}
\end{table}

\subsection{Statistical Testing with Uncertainty Sampling} \label{appendix_subsec:statistical_testing}
To further investigate the effect of Uncertainty Sampling, we perform experimentation comparing this method to the baseline over 20 different seeds. While only using 5 random seeds did not show a statistically significant result $(p<0.05)$, we find that this is a result of the small sample size. When we use a larger sample size we do find statistically significant improvements on the most challenging OOD examples from the Uncertainty Sampling method (see Table \ref{tab:stat_testing_uncertainty_sampling}). 

\begin{table*}
\centering
\resizebox{\textwidth}{!}{
 \begin{tabular}{l c c c c c c c c @{\hskip 0.5em} c @{\hskip 0.5em} c @ {\hskip 0.5em} 
 c c c c}
  \toprule

& & \multicolumn{6}{c}{Challenge-OOD} & & \multicolumn{5}{c}{Standard-OOD}  \\ 
    \cmidrule{3-8} \cmidrule{10-14} 
    & SNLI & r1 & r2 & r3 \text{ } & COPA & INLI-I & WANLI & Avg. & MNLI-m & MNLI-mm & FEVER & Scitail & INLI-NLI & Avg. \\
  \midrule

\rowcolor{light-gray} \multicolumn{15}{c}{Fine-tuning Baselines} \\
Baseline (10k) & 92.45 \text{  } \text{  } & 66.10 & \textbf{58.70} & 55.09 & 56.45 & 60.38 & 60.73 & 59.57\text{  } \text{  } & 86.90 & 87.25 & 70.98 & \textbf{73.32} & 80.56 & 79.80\text{  } \\
   \midrule
  \rowcolor{light-gray} \multicolumn{15}{c}{Sampling Methods} \\
Uncertainty Sampling & \textbf{92.60} $\uparrow$$\dagger$ & \textbf{66.90} & 58.67 & \textbf{55.58} & \textbf{60.89} & \textbf{60.25} & \textbf{60.75} & \textbf{60.51}$\uparrow$$\dagger$ & \textbf{87.56} & \textbf{87.71} & \textbf{71.11} & 72.15 & \textbf{81.01} & \textbf{79.91}$\uparrow$ \\
\bottomrule
  \end{tabular}}
\caption{Accuracy of the GPT-4o-mini baseline compared to the Uncertainty Sampling method using 20 seeds. $\dagger$ represents results where $p < 0.05$, using two-tailed bootstrapping statistical testing \cite{efron1993introduction}. The best results from fine-tuned models are in \textbf{bold}.}
\label{tab:stat_testing_uncertainty_sampling}
\end{table*}

\subsection{Ablation Experiments} \label{appendix_subsec:ablation_experiments}

\paragraph{Testing the if in doubt, discard validation.} \label{sec:appendix_iidd_validation}
We investigate the effect of using \textit{if in doubt, discard} validation to improve the quality of our labelled data. While there is an improvement of 2.26\% for the Long \& Complex Generation on Challenge-OOD (Table \ref{tab:appendix_table_idd}), there is no corresponding improvement on Standard-OOD. We also find little difference in performance on the MNLI Domains Generation, for either Challenge-OOD or Standard-OOD. We conclude that the \textit{if in doubt, discard} validation is helpful for the most complex LLM-generated data, but is otherwise not required.

\begin{table*}
\centering
\resizebox{\textwidth}{!}{
 \begin{tabular}{l c c c c c c c c @{\hskip 0.5em} c @{\hskip 0.5em} c @ {\hskip 0.5em} 
 c c c c}
  \toprule

& & \multicolumn{6}{c}{Challenge-OOD} & & \multicolumn{5}{c}{Standard-OOD}  \\ 
    \cmidrule{3-8} \cmidrule{10-14} 
    & SNLI & r1 & r2 & r3 \text{ } & COPA & INLI-I & WANLI & Avg. & MNLI-m & MNLI-mm & FEVER & Scitail & INLI-NLI & Avg. \\
  \midrule
    \rowcolor{light-gray} \multicolumn{15}{c}{GPT-4o-mini - MNLI Domains Generation} \\
Without validation & 92.48 & \textbf{68.56} & \textbf{58.08} & \textbf{54.18} & 52.96 & 54.92 & \textbf{63.02} & 58.62  & 87.71 & 88.20 & \textbf{73.39} & \textbf{81.23} & \textbf{86.21} & \textbf{83.35} \\ % \\ % & \textbf{73.39} & 87.71 & 88.20 & \textbf{81.23} & \textbf{86.21} & \textbf{83.35} \\
With Validation & \textbf{92.59} & 67.72 & 57.96 & 54.00 & \textbf{54.80} & \textbf{56.04} & 62.89 & \textbf{58.90} & \textbf{87.79} & \textbf{88.34} & 72.97 & 80.98 & 85.90 & 83.20 \\ % & 57.96 & 54.00 & \textbf{54.80} & \textbf{56.04} & 62.89 & \textbf{58.90} \\ % & 72.97 & \textbf{87.79} & \textbf{88.34} & 80.98 & 85.90 & 83.20 \\
  \midrule
  \rowcolor{light-gray} \multicolumn{15}{c}{GPT-4o-mini - Long \& Complex Generation} \\
Without validation & \textbf{92.48}  & 68.52 & 59.74 & 52.22 & \textbf{57.58} & 50.34 & \textbf{63.26} & 58.61  & \textbf{88.02} & \textbf{88.74} & \textbf{73.03} & \textbf{79.67} & 83.34 & \textbf{82.56} \\ % & \textbf{59.74} & 52.22 & \textbf{57.58} & 50.34 & \textbf{63.26} & 58.61 \\ % & \textbf{73.03} & \textbf{88.02} & \textbf{88.74} & \textbf{79.67} & 83.34 & \textbf{82.56} \\
With Validation & 92.26 & \textbf{69.34} & \textbf{60.28} & \textbf{57.42} & 55.90 & \textbf{61.42} & 60.86 & \textbf{60.87} & 87.74 & 88.07 & 72.85 & 76.72 & \textbf{85.83} & 82.24 \\
%With Validation & \textbf{92.64} & 67.24 & 59.04 & \textbf{56.03} & 55.68 & \textbf{61.14} & 60.35 & \textbf{59.91} & 87.56 & 88.03 & 72.06 & 76.20 & \textbf{85.73} & 81.92  \\ % & 59.04 & \textbf{56.03} & 55.68 & \textbf{61.14} & 60.35 & \textbf{59.91} \\ % & 72.06 & 87.56 & 88.03 & 76.20 & \textbf{85.73} & 81.92 \\
\bottomrule
  \end{tabular}}
\caption{We compare results for our data generation methods, showing performance with and without the \textit{if in doubt, discard validation}.}
\label{tab:appendix_table_idd}
\end{table*}

\paragraph{Selecting both $\mathcal{D}_\text{down}$ and $\mathcal{D}_\text{up}$.}

\begin{table*}
\centering
\resizebox{\textwidth}{!}{
 \begin{tabular}{l c c c c c c c c @{\hskip 0.5em} c @{\hskip 0.5em} c @ {\hskip 0.5em} 
 c c c c}
  \toprule

& & \multicolumn{6}{c}{Challenge-OOD} & & \multicolumn{5}{c}{Standard-OOD}  \\ 
    \cmidrule{3-8} \cmidrule{10-14} 
    & SNLI & r1 & r2 & r3 \text{ } & COPA & INLI-I & WANLI & Avg. & MNLI-m & MNLI-mm & FEVER & Scitail & INLI-NLI & Avg. \\
  \midrule
  \rowcolor{light-gray} \multicolumn{15}{c}{GPT-4o-mini - Uncertainty Sampling} \\
Up selection & 92.80 & \textbf{67.42} & 58.60 & \textbf{55.12} & \textbf{63.68} & 60.60 & \textbf{60.99} & \textbf{61.07} \text{ }& \textbf{88.14} & \textbf{88.24} & 71.57 & 71.47 & 82.01 & 80.28 \text{ } \\ %& \textbf{67.42} & 58.60 & \textbf{55.12} & \textbf{63.68} & 60.60 & \textbf{60.99} & \textbf{61.07} \text{ } \\ % & 71.57 & \textbf{88.14} & \textbf{88.24} & 71.47 & 82.01 & 80.29 \text{ }
Up and down & \textbf{92.83} & 67.16 & \textbf{59.54} & 54.97 & 60.14 & \textbf{60.74} & 61.51 & 60.68\text{ } & 87.93 & 88.11 & \textbf{71.46} & \textbf{71.93} & \textbf{82.33} & \textbf{80.35}$\uparrow$ \\ % & 67.26 & \textbf{59.54} & 54.97 & 60.14 & \textbf{60.7}4 & 61.51 & 60.68$\downarrow$ \\ % & \textbf{72.46} & 87.93 & 88.11 & 71.93 & 82.33 & \textbf{80.35}$\uparrow$ \\
  \midrule
  \rowcolor{light-gray} \multicolumn{15}{c}{GPT-4o-mini - Difficulty Score Sampling} \\
Up selection & 92.66 & \textbf{67.48} & \textbf{59.10} & \textbf{56.42} & \textbf{58.84} & 59.54 & \textbf{61.40} & 60.46 & \textbf{87.64} & \textbf{87.66} & \textbf{71.61} & \textbf{71.88} & 80.81 & \textbf{79.92} \\

Up and down & \textbf{92.68}  & 66.86 & 57.96 & 56.35 & 57.58 & \textbf{64.94} & 60.39 & \textbf{60.68} & 86.77 & 86.96 & 71.18 & 71.38 & \textbf{81.12} & 79.48 \\ % & 66.86 & 57.96 & 56.35 & 57.58 & \textbf{64.94} & 60.39 & \textbf{60.68} \\
\bottomrule
  \end{tabular}}
\caption{We experiment with selecting new examples for $\mathcal{D}_\text{up}$ using our Uncertainty Sampling, while also choosing examples for $\mathcal{D}_\text{down}$ as the most confident predictions within  $\mathcal{D}_\text{init}$. `Up selection' refers to the Uncertainty Sampling method defined in the paper, where we choose examples based on their uncertainty, and randomly choose examples in $\mathcal{D}_\text{init}$ for $\mathcal{D}_\text{down}$. `Up and down' also chooses $\mathcal{D}_\text{down}$ based on model confidence. We also perform a similar experiment with our Difficulty Score Sampling method, choosing $\mathcal{D}_\text{down}$ based on the lowest difficulty scores.}
\label{tab:up_and_down_table}
\end{table*}

We experiment with using our sampling methods to choose both $\mathcal{D}_\text{down}$ and $\mathcal{D}_\text{up}$, using both the Uncertainty Sampling and Difficulty Score Sampling methods. For Uncertainty Sampling, we choose $\mathcal{D}_\text{up}$ as the examples with the highest entropy in $\mathcal{D}_\text{potential}$, while choosing $\mathcal{D}_\text{down}$ as the examples with the lowest entropy in $\mathcal{D}_\text{init}$. Similarly, for Difficulty Score Sampling, we choose $\mathcal{D}_\text{down}$ as the examples in $\mathcal{D}_\text{init}$ with the lowest difficulty score.

We find that using our methods to also select examples for $\mathcal{D}_\text{down}$ makes little difference to performance (Table \ref{tab:appendix_uncertainty_sampling_tables}). This suggests that additionally including a small number of complex examples has a much greater effect than removing a small number of easier examples from the model's training data. %Based on these results, we do not further investigate strategies for choosing $\mathcal{D}_{down}$.

%\section{Results when training with MNLI} %\label{sec:appendix_train_mnli_results}
%\input{latex/Tables/mnli_results_table}

\paragraph{Testing Difficulty Score Sampling variations.}  \label{sec:appendix_difficulty_score_sampling}
\begin{table*}
\centering
\resizebox{\textwidth}{!}{
   \begin{tabular}{l c c c c c c c c @{\hskip 0.5em} c @{\hskip 0.5em} c @ {\hskip 0.5em} 
 c c c c}
  \toprule
& & \multicolumn{6}{c}{Challenge-OOD} & & \multicolumn{5}{c}{Standard-OOD}  \\ 
    \cmidrule{3-8} \cmidrule{10-14} 
    & SNLI & r1 & r2 & r3 \text{ } & COPA & INLI-I & WANLI & Avg. & MNLI-m & MNLI-mm & FEVER & Scitail & INLI-NLI & Avg. \\
  \midrule
  \rowcolor{light-gray} \multicolumn{15}{c}{GPT-4o-mini - Difficulty Score Sampling} \\
C + D  & 92.55 & 67.42 & 58.34 & 54.98 & 57.58 & \textbf{59.54} & 61.08 & 59.82 & 87.93 & 87.79 & 71.59 & 70.62 & 80.61 & 79.71 \\ % & 67.42 & 58.34 & 54.98 & 57.58 & 59.54 & 61.08 & 59.82 \\ % & 71.59 & 87.93 & 87.79 & 70.62 & 80.61 & 79.71 \\
D & \textbf{92.66} & 67.48 & \textbf{59.10} & \textbf{56.42} & 58.84 & \textbf{59.54} & \textbf{61.40} & \textbf{60.46} & 87.64 & 87.66 & \textbf{71.61} & 71.88 & 80.81 & 79.92 \\
C + D + P + F & 92.41  & \textbf{67.92} & 57.94 & 55.50 & \textbf{59.64} & 56.32 & 61.07 & 59.73 & \textbf{87.94} & \textbf{87.82} & 71.35 & \textbf{72.39} & \textbf{80.93} & \textbf{80.09} \\ % & \textbf{67.92} & 57.94 & 55.50 & \textbf{59.64} & 56.32 & 61.07 & 59.73 \\ % & 71.35 & 87.94 & 87.82 & 72.39 & 80.93 & 80.09 \\ 

\bottomrule
  \end{tabular}}
\caption{We experiment with different methods for combining the scores provided by the baseline model $\mathcal{M}$ before fine-tuning. These scores are identified by asking the model to make few-shot assessments about the examples in $\mathcal{D}_\text{potential}$. The available scores are: C (correctness), D (difficulty), P (plausibility), F (fluency). C + D is the configuration used in the main paper.}
\label{tab:appendix_qa_tables}
\end{table*}

We investigate introducing a wider range of scores to measure the quality of NLI training examples, beyond the difficulty score used within our Difficulty Score Sampling. We try two different configurations, firstly including a score for label quality and summing this together with the difficulty score, and secondly, also adding scores for the plausibility and fluency of the instances (see Table \ref{tab:appendix_qa_tables}).

While using the combination of all of the scores (difficulty, correctness, plausibility and fluency) performs slightly better on Standard-OOD, performance is worse on Challenge-OOD. We conclude that including the additional scores does not help to improve model robustness.

\paragraph{Testing Uncertainty Sampling variations} \label{sec:appendix_uncertainty_sampling}
\begin{table*}
\centering
\resizebox{\textwidth}{!}{
 \begin{tabular}{l c c c c c c c c @{\hskip 0.5em} c @{\hskip 0.5em} c @ {\hskip 0.5em} 
 c c c c}
  \toprule
& & \multicolumn{6}{c}{Challenge-OOD} & & \multicolumn{5}{c}{Standard-OOD}  \\ 
    \cmidrule{3-8} \cmidrule{10-14} 
    & SNLI & r1 & r2 & r3 \text{ } & COPA & INLI-I & WANLI & Avg. & MNLI-m & MNLI-mm & FEVER & Scitail & INLI-NLI & Avg. \\
  \midrule
  \rowcolor{light-gray} \multicolumn{15}{c}{GPT-4o-mini - Uncertainty Sampling} \\
Uncertainty Sampling & \textbf{92.80} & \textbf{67.42} &\textbf{58.60} & \textbf{55.12} & \textbf{63.68} & \textbf{60.60} & 60.99 & \textbf{61.07}  & 88.14 & \textbf{88.24} & 71.57 & 71.47 & \textbf{82.01} & 80.28  \\ % &\textbf{58.60} & \textbf{55.12} & \textbf{63.68} & \textbf{60.60} & \textbf{60.99} & \textbf{61.07} \\ % & 71.57 & 88.14 & \textbf{88.24} & 71.47 & \textbf{82.01} & 80.29 \\
Uncertainty Sampling (correct only) & 92.68 & 66.90 & 57.52 & 54.88 & 57.10 & 59.06 & \textbf{61.62} & 59.51 & \textbf{88.21} & \textbf{88.24} & \textbf{71.77} & \textbf{75.39} & 80.88 & \textbf{80.90} \\ % & 57.52 & 54.88 & 57.10 & 59.06 & 61.62 & 59.51 \\ % & \textbf{71.77} & \textbf{88.21} & \textbf{88.24} & \textbf{75.39} & 80.88 & \textbf{80.90} \\

\bottomrule
  \end{tabular}}
\caption{We experiment with only choosing examples for $\mathcal{D}_{up}$ for Uncertainty Sampling if the predictions from the baseline model $\mathcal{M}_{base}$ are correct.}
\label{tab:appendix_uncertainty_sampling_tables}
\end{table*}

With the aim of reducing label noise with our Uncertainty Sampling method, we experiment with only choosing examples for $\mathcal{D}_\text{up}$ which are correctly predicted by $\mathcal{M}$ (see Table \ref{tab:appendix_uncertainty_sampling_tables}). This means we choose examples where the model is correct, but still uncertain. We find however that this restriction results in worse robustness on Challenge-OOD, presumably because the selected examples are now less complex. We do however find a small improvement in Standard-OOD performance.

\paragraph{Replacing more training examples when sampling.} \label{sec:appendix_replacing_more_training_examples}
The experimentation in this work considers the value of $\mathcal{K}$ as 5\% of the training sample size, with each experiment in the paper replacing $\mathcal{K}$ many instances of each class (changing 15\% of the training sample in total).
In this section, we experiment with increasing this value of $\mathcal{K}$ to measure whether replacing more training instances leads to further improvements in robustness. These experiments use the Uncertainty Sampling method, which constructs $\mathcal{D}_\text{up}$ with the $\mathcal{K}$ most uncertain instances for each class based on the predictions from $\mathcal{M}_\text{base}$.

We find that increasing the size of $\mathcal{K}$ leads to a small reduction in performance both in-distribution and on Challenge-OOD (see Table \ref{tab:appendix_changing_value_of_K}). This suggests that while more complex training examples can help improve the model's performance on challenging out-of-distribution examples, if too many of the training examples are replaced then the challenging examples may become over-represented.

\begin{table*}
\centering
\resizebox{\textwidth}{!}{
 \begin{tabular}{l c c c c c c c c @{\hskip 0.5em} c @{\hskip 0.5em} c @ {\hskip 0.5em} 
 c c c c}
  \toprule
& & \multicolumn{6}{c}{Challenge-OOD} & & \multicolumn{5}{c}{Standard-OOD}  \\ 
    \cmidrule{3-8} \cmidrule{10-14} 
    & SNLI & r1 & r2 & r3 \text{ } & COPA & INLI-I & WANLI & Avg. & MNLI-m & MNLI-mm & FEVER & Scitail & INLI-NLI & Avg. \\
  \midrule
  \rowcolor{light-gray} \multicolumn{15}{c}{GPT-4o-mini - Uncertainty Sampling} \\
K = 0\% ($\mathcal{M}_{base}$ baseline) & 92.47 & 65.98 & 58.10 & \textbf{55.38} & 56.20 & \textbf{61.42} & 60.62 & 59.62 & 86.93 & 87.24 & 71.21 & 72.55 &  80.21 & 79.63 \\
K = 5\% (main results) & \textbf{92.80} & \textbf{67.42} & \textbf{58.60} & 55.12 & \textbf{63.68} & 60.60 & 60.99 & \textbf{61.07}  & 88.14 & 88.24 & 71.57 & 71.47 & \textbf{82.01} & 80.28  \\
K = 10\% & 92.73 & 67.32 & 57.54 & 53.93 & 58.98 & 55.94 & 62.76 & 59.41 & 88.47 & 88.76 & 72.02 & 74.62 & 81.67 & 81.11 \\
K = 15\% & 92.61 & 68.10 & 57.52 & 53.38 & 61.54 & 57.62 & \textbf{62.88} & 60.21 & \textbf{88.62} & \textbf{88.80} & \textbf{72.54} & \textbf{75.36} & 81.24 & \textbf{81.31} \\
\bottomrule
  \end{tabular}}
\caption{We experiment with changing the value of $\mathcal{K}$ to measure changes in performance (in-distribution and out-of-distribution) when modifying more of the original training sample. These experiments use the Uncertainty Sampling method.}
\label{tab:appendix_changing_value_of_K}
\end{table*}

\paragraph{Different categorisation of datasets into Standard-OOD and Challenge-OOD.} 
The categorisation of datasets into Standard-OOD and Challenge-OOD is based on the performance of the baseline GPT-4o-mini model. Datasets where this model achieves 70\% accuracy or more are categorised as Standard-OOD, while datasets with worse performance are categorised as Challenge-OOD. Table \ref{tab:appendix_different_standard_challenge_categorisations} shows the results for Standard-OOD and Challenge-OOD with different categorisations of the datasets, either including FEVER in Challenge-OOD or including INLI-I in Standard-OOD. %FEVER is the dataset with the lowest performance in Standard-OOD, while INLI-I is the dataset with the highest performance in Standard-OOD (excluding the r1 ANLI test set, which we keep together with the r2 and r3 test sets).

In each of the different categorisations, we find that the sampling methods perform better on Challenge-OOD compared to the baseline. We also find that the generation methods consistently outperform the baseline for Standard-OOD. In each case, the Long \& Complex Generation outperforms the baseline for both Challenge-OOD and Standard-OOD. 

\begin{table*}
\centering
\resizebox{\textwidth}{!}{
 \begin{tabular}{l c c c c c c c}
  \toprule

& \multicolumn{3}{c}{Challenge-OOD} & & \multicolumn{3}{c}{Standard-OOD}  \\ 
    \cmidrule{2-4} \cmidrule{6-8} 
    & Standard & Fever in Chal-OOD & INLI-I in Std-OOD & & Standard & Fever in Chal-OOD & INLI-I in Std-OOD \\
  \midrule
%  \multicolumn{15}{c}
%{\textbf{LLM with no FT:}} \\
%LLM w/o FT & 85.02 \text{   } \text{   } & 77.00 & 63.90 & 58.33 & 63.60 & 60.80 & 64.88 & 64.75 \text{   } & 84.62 & 85.09 & 73.53 & 84.62 & 87.30 & 83.03 \text{   } \text{   } \\
%\midrule

\rowcolor{light-gray}\multicolumn{8}{c}{Fine-tuning Baselines} \\

Baseline (10k) & 59.62 & 61.27 & 59.26 & & 79.63 & 81.73 & 76.59 \\% & 65.98 & 58.10 & 55.38 & 56.20 & 61.42 & 60.62 & 59.62 \text{ } \\ %& 71.21 & 86.93 & 87.24 & 72.55 & 80.21 & 79.63 \text{  }\\
% &  &  &  &  &  &  &  & - & & &  & & & - \\
%Random Sampling & 59.51 & & & & 79.16 \\ % & 65.80 & 58.66 & 55.37 & 56.48 & 60.60 & 60.15 & 59.51$\downarrow$ \\% & 71.17 & 86.94 & 87.00 & 69.56 & 81.15 & 79.16 $\downarrow$ \\
% & +0.08 & (0.18) & +0.56 & (0.01) & +0.28 & (0.82) & (0.46) & (0.11) & (0.04) & +0.01 & (0.24) & (2.99) & +0.94 & -0.46 \\
% &  &  &  &  &  &  &  & (0.11) & & &  & & & (0.46) \\
   \midrule

% & (0.15) & (0.50) & (0.12) & (2.65) & +6.75 & (4.54) & (0.30) & (0.23) & +0.64 & (0.06) & (0.18) & (5.03) & (0.41) & (1.01) \\
% &  & &  &  & &  & & (0.23) & &  &  &  &  & (1.01) \\
  \rowcolor{light-gray} \multicolumn{8}{c}{Sampling Methods} \\
Misclassified Sampling & 59.39 & 61.17 & 59.89 & & 78.62 & 80.31 & 75.00 \\ %& 65.48 & 57.98 & 52.73 & 62.94 & 56.88 & 60.32 & 59.39$\downarrow$\\% & 71.85 & 86.87 & 87.06 & 67.52 & 79.80 & 78.62 $\downarrow$ \\

% Correct \& Uncertain & 92.68 & 66.90 & 57.52 & 54.88 & 57.10 & 59.06 & \textbf{61.62} & 59.51$\downarrow$ & 71.74 & 88.21 & 88.24 & 75.39 & 80.88 & 80.89 $\uparrow$\\
% &  &  &  &  &  &  &  & (0.10) & &  &  & & & +1.26 \\
% & +0.08 & +1.44 & +0.24 & (0.40) & +1.38 & (1.88) & +0.46 & +0.21 & +0.38 & +0.86 & +0.69 & (1.93) & +0.40 & +0.08 \\
% &  &  & &  &  &  &  & +0.21 & & & & & & +0.08 \\
Concat Hypothesis Sampling & 60.77 & 62.20 & 60.52 & & 78.93 & 80.96 & 76.11 \\ % & 65.72 & 58.14 & 55.88 & 62.84 & \textbf{62.04} & 60.00 & 60.77$\uparrow$ \\ % & 70.81 & 86.55 & 86.69 & 69.88 & 80.71 & 78.93 $\downarrow$\\
% & +0.09 & (0.26) & +0.04 & +0.50 & +6.64 & +0.02 & (0.62) & +1.05 & (0.40) & (0.38) & (0.55) & (2.67)  & +0.50 & (0.70) \\
% & &  & &  &  &  &  & +1.05 & &  &  &   & & (0.70) \\
Difficulty Score Sampling & 60.46 & 61.94 & 60.65 & & 79.92 & 80.96 & 75.70 \\
%79.92 \\ % & 67.48 & 59.10 & 56.42 & 58.84 & 59.54 & 61.40 & 60.46$\uparrow$ \\ % & 71.61 & 87.64 & 87.66 & 71.88 & 80.81 & 79.92 $\uparrow$
Uncertainty Sampling & \textbf{61.07} & 62.57 & \textbf{61.16} & & 80.28 & 82.47 & 77.01 \\% & 67.42 & 58.60 & 55.12 & \textbf{63.68} & 60.60 & 60.99 & \textbf{61.07}$\uparrow$ \\ % & 71.57 & \textbf{88.14} & 88.24 & 71.47 & 82.01 & 80.28 $\uparrow$ \\
% & +0.33 & +1.44 & +0.50 & (0.26) & +7.48 & (0.82) & +0.37 & +1.45 & +0.36 & +1.21 & +1.00 & (1.08) & +1.80 & +0.66 \\
% &  &  &  & &  &  &  & \textbf{+1.45} &  &  &  &  &  & +0.66 \\
 \midrule
   \rowcolor{light-gray} \multicolumn{8}{c}{Generated Data} \\

MNLI Domains Generation & 58.90 & 60.91 & 59.47 & & 83.20 & \textbf{85.75} & 78.67 \\ % & 67.72 & 57.96 & 54.00 & 54.80 & 56.04 & 62.89 & 58.90$\downarrow$ \\ % & 72.97 & 87.79 & 88.34 & 80.98 & 85.90 & 83.20 $\uparrow$ \\

Short \& Simple Generation & 58.81 & 60.91 & 60.17 & & \textbf{83.29} & 85.74 & 78.08 \\ %& 68.88 & 58.14 & 53.43 & 56.28 & 52.02 & \textbf{64.11} & 58.81$\downarrow$ \\ % & 73.48 & 87.88 & 88.66 & \textbf{81.67} & 84.76 & \textbf{83.29} $\uparrow$ \\

Long \& Simple Generation & 58.43 & 60.52 & 60.11 & & 82.46 & 84.82 & 77.05 \\ %& \textbf{69.62} & 60.40 & 52.58 & 55.22 & 50.00 & 62.75 & 58.43$\downarrow$ \\ %& 73.05 & 88.08 & \textbf{88.88} & 78.97 & 83.33 & 82.46 $\uparrow$ \\

Long \& Complex Generation & 60.87 & \textbf{62.58} & 60.76 & & 82.24 & 84.59 & \textbf{78.77} \\ %& 67.24 & \textbf{59.04} & \textbf{56.03} & 55.68 & 61.14 & 60.35 & 59.91$\uparrow$ \\ %& \textbf{72.06} & 87.56 & 88.03 & 76.20 & \textbf{85.73} & 81.92 $\uparrow$ \\
% & +0.17 & +1.26 & +0.94 & +0.65 & (0.52) & (0.28) & (0.27) & +0.30 & +0.85 & +0.63 & +0.79 & +3.65 & +5.52 & +2.29 
% &  &  &  &  &  &  &  & +0.30 &  &  &  &  & & \textbf{+2.29} 
\bottomrule
  \end{tabular}}
\caption{Changing the categorisation of datasets to Standard-OOD and Challenge-OOD, either including Fever in Challenge-OOD or including INLI-I in Standard-OOD. Our findings are consistent across these categorisations.}
\label{tab:appendix_different_standard_challenge_categorisations}
\end{table*}

\section{Maths Datasets} \label{sec:appendix_maths_experiments}
\begin{table*}
\centering
\resizebox{\textwidth}{!}{
  \begin{tabular}{l @{\hskip 0.5em} c c @{\hskip 0.5em} c @{\hskip 0.5em} c c c c c c}
  \toprule

& \multicolumn{1}{c}{ID} & & \multicolumn{5}{c}{OOD}  \\ 
    \cmidrule{3-9} 
    & Calc-Ape210k & AQuA-RAT & MinervaMath & MATH500 & Omni-MATH500 & AMC23 & OlympiadBench & HMMT & Avg.\\
  \midrule
Few-shot & 29.02 & 30.73 & 14.66 & 22.73 & 12.46 & 20.00 & 9.31 & 1.14 & 15.86 \\
  \midrule
10k baseline & 49.05 & 32.11 & 36.44 & \textbf{23.74} & 13.84 & 16.50 & 10.02 & 1.14 & 19.11 \\
Uncertainty Sampling ($\mathcal{K}$ = 15\%) & \textbf{49.24} & \textbf{33.67} & 36.65 & 23.33 & 14.25 & 14.50 & \textbf{10.61} & \textbf{1.81} & 19.26 \\
Uncertainty Sampling ($\mathcal{K}$ = 50\%) & 48.74 & 32.93 & \textbf{37.70} & 23.08 & \textbf{14.58} & \textbf{19.00} & 10.14 & 1.59 & \textbf{19.86} \\
\bottomrule
  \end{tabular}}
\caption{We test our Uncertainty Sampling method on maths datasets after training on 10k instances from Calc-Ape, before testing performance on other out-of-distribution maths datasets. We experiment with selecting $\mathcal{K}$ = |$\mathcal{D}_{up}$| = 0.15 $\times$ |$\mathcal{D}_{init}$|, similar to our NLI experiments, and $\mathcal{K}$ = |$\mathcal{D}_{up}$| = 0.50 $\times$ |$\mathcal{D}_{init}$|.}
\label{tab:appendix_maths_experiments}
\end{table*}
As our Uncertainty Sampling method is directly applicable to a range of other NLP tasks, we experiment with applying this method to maths datasets. We train our model on 10,000 instances from Calc-Ape210k \cite{kadlcik-etal-2023-soft}, before evaluating on AQuA-RAT \cite{ling2017program}, MinervaMATH\cite{10.5555/3600270.3600548}, MATH500 \cite{lightman2024lets}, Omni-MATH500 \cite{gao2025omnimath}, AMC23 \cite{yang2024qwen25mathtechnicalreportmathematical}, OlympiadBench \cite{he-etal-2024-olympiadbench} and HMMT \cite{smolagents}. Similar to our experiments with NLI, we aim to evaluate out-of-distribution performance on a wide range of datasets. 

For AQuA-RAT, we predict numerical answers rather than choosing multiple choice labels (also without providing the model with the multiple choice answers in the prompt). We also remove questions requiring ``All of the above'' or ``None of the above'' as answers, reducing the test set from 254 to 218 samples. For MinervaMATH, MATH500, Omni-MATH500, OlympiadBench and HMMT we remove questions with answers that are not a number. This reduces the test sets to 191, 396, 321, 505 and 88 instances respectively. For HMMT, this involves merging HMMT Feb. 23, HMMT Feb 24 and HMMT Feb. 25. To be consistent with our NLI experiments, and to avoid larger fine-tuning costs, we do not train our model to generate chain-of-thought answers.

We experiment with $\mathcal{K}$ = |$\mathcal{D}_\text{up}$| = 0.15 $\times$ |$\mathcal{D}_\text{init}$|, similar to our NLI experiments. But unlike NLI, there is less variety in the complexity of the different training examples in Calc-Ape210k, so we also experiment with $\mathcal{K}$ = |$\mathcal{D}_\text{up}$| = 0.50 $\times$ |$\mathcal{D}_\text{init}$|. 
We find that when we set $\mathcal{K}$ = |$\mathcal{D}_\text{up}$| = 0.50, Uncertainty Sampling improves out-of-distribution performance by 0.75 percentage points (from 19.11\% to 19.86\%), an increase of 3.9\%.

\section{Further Details on Data Sampling, Generation, and Filtering}\label{appendix_sec:further_details}

\paragraph{Generating Complex Hypotheses}
When generating synthetic data with more complex entailment relationships, we use the following prompts to encourage hypotheses that are related to multiple parts of the premise input:

\begin{itemize}
    %\itemsep-2pt
    %\itemindent-10pt
    \item \textbf{\textit{Contradiction Strategy 1}}: "From the premise above, randomly pick two sentences, label them sentence 1 and sentence 2. Then construct a sentence that contradicts sentence 1, does not contradict sentence 2, but mentions something related to sentence 2"
    \item \textbf{\textit{Contradiction Strategy 2}}: "From the premise above, randomly pick two sentences, label them sentence 1 and sentence 2. Then construct a two sentence output which contains one sentence implied by sentences 1 and 2, and another sentence that contradicts one of them."
    \item \textbf{\textit{Neutral Strategy 1:}} "From the premise above, randomly pick two sentences, label them sentence 1 and sentence 2. Then construct a sentence that relates to both sentences 1 and 2, but is not implied by them."
    \item \textbf{\textit{Neutral Strategy 2:}} "From the premise above, randomly pick two sentences, label them sentence 1 and sentence 2. Then construct a two sentence output which contains one sentence implied by sentences 1 and 2, and another sentence that is not implied by them."
    \item \textbf{\textit{Entailment Strategy 1:}} "From the premise above, randomly pick two sentences, label them sentence 1 and sentence 2. Then construct a sentence that is implied by both sentences 1 and 2, relating to both of these."
    \item \textbf{\textit{Entailment Strategy 2:}} "Provide a very short, concise fact that is strictly implied by a specific part of the premise above. Don't do something obvious."
\end{itemize}

\paragraph{Unlabelled Data Additional Filtering.}

When using the unlabelled data for the MNLI training domains provided by \citet{stacey-rei-2024-distilling}, we observe a number of cases where multiple premise and hypotheses have been generated by an LLM for a single instance. We filter out instances that contain the words `premise', `hypothesis', `entailment', `neutral', `contradiction', `implies' or `implied'. We also filter out premises with `sure!', `can i help', `happy to help' and `no problem'. We further filter examples ending in a question mark or exclamation mark, and where two sentences have been combined without correct punctuation.

\paragraph{Unlabelled Data Domain Information} \label{sec:Unlab_Data_Domains}
When using our LLMs to generate additional data for 51 domains, we ask the LLM to either `Provide a single sentence' or `Provide an output with four sentences' about: `about a workplace', `about the founding fathers', `from a review of a book', `overheard from radio podcast', `that you might say to a tourist when recommending a specific location', `describing your favourite place in the city', `that you would never want to hear', `that you might overhear on public transport', `about someone you admire', `about someone historically significant', `from a boring website', `to explain a complicated concept', `that you only hear in films', `from a not so famous speech', `that you might read in a blog about geography', `that might surprise me', `you remember from when someone read you the riot act', `from some song lyrics', `extract from a made-up newspaper', `describing a musician', `that you might hear in a battle', `from a propaganda leaflet', `with a fact that surprised you', `from a Shakespeare play', `containing a joke', `extract from Poirot', `explaining investment banking to a graduate student', `describing the history of mathematics', `about railways', `that you might read in a formal legal contract', `describing a medical procedure', `that you could hear in an airplane cabin', `giving orders on a submarine', `with technical analysis', `describing the plants grown in a garden', `explaining an animal found in the rainforest', `found in the instructions for a electrical item', `from a poem', 'from a detailed a technical report', `in slang', ``describing a tactical analysis of the team's strategy", `overheard between two people in a pub', `from a review of an NLP paper submitted for peer review', `with a surprising historical fact', `that you could have overheard in the Great Exhibition', 'about farming', `that you could hear in an undergraduate lecture', `describing the properties of an element in the periodic table', `that might be interesting to an economist', `describing a scientific advancement' or `to describe an object in the British Museum'.

\paragraph{Method Standard Deviations.} \label{sec:appendix_std_dev}
\begin{table*}
\centering
\resizebox{\textwidth}{!}{
 \begin{tabular}{l c c c c c c c c @{\hskip 0.5em} c @{\hskip 0.5em} c @ {\hskip 0.5em} 
 c c c c}
  \toprule
& & \multicolumn{6}{c}{Challenge-OOD} & & \multicolumn{5}{c}{Standard-OOD}  \\ 
    \cmidrule{3-8} \cmidrule{10-14} 
    & SNLI & r1 & r2 & r3 \text{ } & COPA & INLI-I & WANLI & \text{   } & MNLI-m & MNLI-mm & FEVER & Scitail & INLI-NLI & \text{   } \\
  \midrule
  \rowcolor{light-gray} \multicolumn{15}{c}{Baselines} \\
Baseline (10k) & 0.001 & 0.008 & 0.011 & 0.009 & 0.023 & 0.043 & 0.007 & &  0.005 & 0.007 & 0.007 & 0.034 & 0.012 \\% & 65.98 & 58.10 & 55.38 & 56.20 & 61.42 & 60.62 & 59.62 \text{ } \\ %& 71.21 & 86.93 & 87.24 & 72.55 & 80.21 & 79.63 \text{  }\\
% &  &  &  &  &  &  &  & - & & &  & & & - \\
Random sampling & 0.001 & 0.007 & 0.014 & 0.017 & 0.047 & 0.053 & 0.017 & & 0.009 & 0.011 & 0.010 & 0.025 & 0.007 \\ % & 65.80 & 58.66 & 55.37 & 56.48 & 60.60 & 60.15 & 59.51$\downarrow$ \\% & 71.17 & 86.94 & 87.00 & 69.56 & 81.15 & 79.16 $\downarrow$ \\
% & +0.08 & (0.18) & +0.56 & (0.01) & +0.28 & (0.82) & (0.46) & (0.11) & (0.04) & +0.01 & (0.24) & (2.99) & +0.94 & -0.46 \\
% &  &  &  &  &  &  &  & (0.11) & & &  & & & (0.46) \\
   \midrule

% & (0.15) & (0.50) & (0.12) & (2.65) & +6.75 & (4.54) & (0.30) & (0.23) & +0.64 & (0.06) & (0.18) & (5.03) & (0.41) & (1.01) \\
% &  & &  &  & &  & & (0.23) & &  &  &  &  & (1.01) \\
 \rowcolor{light-gray}  \multicolumn{15}{c}{Sampling} \\
Misclassified Sampling & 0.002 & 0.012 & 0.008 & 0.022 & 0.077 & 0.067 & 0.014 & & 0.007 & 0.009 & 0.007 & 0.017 & 0.010 \\ %& 65.48 & 57.98 & 52.73 & 62.94 & 56.88 & 60.32 & 59.39$\downarrow$\\% & 71.85 & 86.87 & 87.06 & 67.52 & 79.80 & 78.62 $\downarrow$ \\

% Correct \& Uncertain & 92.68 & 66.90 & 57.52 & 54.88 & 57.10 & 59.06 & \textbf{61.62} & 59.51$\downarrow$ & 71.74 & 88.21 & 88.24 & 75.39 & 80.88 & 80.89 $\uparrow$\\
% &  &  &  &  &  &  &  & (0.10) & &  &  & & & +1.26 \\
% & +0.08 & +1.44 & +0.24 & (0.40) & +1.38 & (1.88) & +0.46 & +0.21 & +0.38 & +0.86 & +0.69 & (1.93) & +0.40 & +0.08 \\
% &  &  & &  &  &  &  & +0.21 & & & & & & +0.08 \\
Hypothesis Concat Sampling & 0.001 & 0.017 & 0.011 & 0.011 & 0.075 & 0.041 & 0.009 & & 0.007 & 0.006 & 0.008 & 0.024 & 0.009 \\ % & 65.72 & 58.14 & 55.88 & 62.84 & \textbf{62.04} & 60.00 & 60.77$\uparrow$ \\ % & 70.81 & 86.55 & 86.69 & 69.88 & 80.71 & 78.93 $\downarrow$\\
% & +0.09 & (0.26) & +0.04 & +0.50 & +6.64 & +0.02 & (0.62) & +1.05 & (0.40) & (0.38) & (0.55) & (2.67)  & +0.50 & (0.70) \\
% & &  & &  &  &  &  & +1.05 & &  &  &   & & (0.70) \\
Difficulty Score Sampling & 0.002 & 0.011 & 0.015 & 0.023 & 0.031 & 0.054 & 0.014 & & 0.006 & 0.007 & 0.006 & 0.027 & 0.011 \\ % & 67.48 & 59.10 & 56.42 & 58.84 & 59.54 & 61.40 & 60.46$\uparrow$ \\ % & 71.61 & 87.64 & 87.66 & 71.88 & 80.81 & 79.92 $\uparrow$
Uncertainty Sampling &  0.002 & 0.015 & 0.016 & 0.016 & 0.055 & 0.042 & 0.009 & & 0.006 & 0.004 & 0.005 & 0.027 & 0.009 \\% & 67.42 & 58.60 & 55.12 & \textbf{63.68} & 60.60 & 60.99 & \textbf{61.07}$\uparrow$ \\ % & 71.57 & \textbf{88.14} & 88.24 & 71.47 & 82.01 & 80.28 $\uparrow$ \\
% & +0.33 & +1.44 & +0.50 & (0.26) & +7.48 & (0.82) & +0.37 & +1.45 & +0.36 & +1.21 & +1.00 & (1.08) & +1.80 & +0.66 \\
% &  &  &  & &  &  &  & \textbf{+1.45} &  &  &  &  &  & +0.66 \\
 \midrule
   \rowcolor{light-gray} \multicolumn{15}{c}{Generated Data} \\

MNLI Domains Generation & 0.002 & 0.018 & 0.009 & 0.017 & 0.013 & 0.030 & 0.008 & & 0.005 & 0.003 & 0.005 & 0.011 & 0.030 \\ % & 67.72 & 57.96 & 54.00 & 54.80 & 56.04 & 62.89 & 58.90$\downarrow$ \\ % & 72.97 & 87.79 & 88.34 & 80.98 & 85.90 & 83.20 $\uparrow$ \\

Short \& Simple Generation & 0.001 & 0.015 & 0.026 & 0.027 & 0.025 & 0.046 & 0.014 & & 0.004 & 0.002 & 0.010 & 0.019 & 0.017 \\ %& 68.88 & 58.14 & 53.43 & 56.28 & 52.02 & \textbf{64.11} & 58.81$\downarrow$ \\ % & 73.48 & 87.88 & 88.66 & \textbf{81.67} & 84.76 & \textbf{83.29} $\uparrow$ \\

Long \& Simple Generation & 0.002 & 0.011 & 0.013 & 0.011 & 0.034 & 0.029 & 0.016 & & 0.001 & 0.003 & 0.003 & 0.016 & 0.017 \\ %& \textbf{69.62} & 60.40 & 52.58 & 55.22 & 50.00 & 62.75 & 58.43$\downarrow$ \\ %& 73.05 & 88.08 & \textbf{88.88} & 78.97 & 83.33 & 82.46 $\uparrow$ \\

Long \& Complex Generation & 0.003 & 0.011 & 0.008 & 0.016 & 0.012 & 0.016 & 0.006 && 0.004 & 0.004 & 0.004 & 0.029 & 0.008 \\ %& 67.24 & \textbf{59.04} & \textbf{56.03} & 55.68 & 61.14 & 60.35 & 59.91$\uparrow$ \\ %& \textbf{72.06} & 87.56 & 88.03 & 76.20 & \textbf{85.73} & 81.92 $\uparrow$ \\
% & +0.17 & +1.26 & +0.94 & +0.65 & (0.52) & (0.28) & (0.27) & +0.30 & +0.85 & +0.63 & +0.79 & +3.65 & +5.52 & +2.29 
% &  &  &  &  &  &  &  & +0.30 &  &  &  &  & & \textbf{+2.29} 
\bottomrule
  \end{tabular}}
\caption{Standard deviations for our results in Table \ref{tab:main_sampling_results}. Standard deviations are from 5 seeds.}
\label{tab:main_results_sd}
\end{table*}

We provide all of the standard deviations for Table \ref{tab:main_sampling_results} in Table \ref{tab:main_results_sd}. As expected, the variance is often large for out-of-distribution test sets \cite{mccoy-etal-2020-berts}. We therefore use 5 random seeds for every experiment (in both the main paper and the appendix), and we compare average performance across multiple different OOD datasets (using our average scores for the Challenge-OOD and Standard-OOD datasets).

\begin{figure*}
\begin{center}
    \includegraphics[width=460pt]{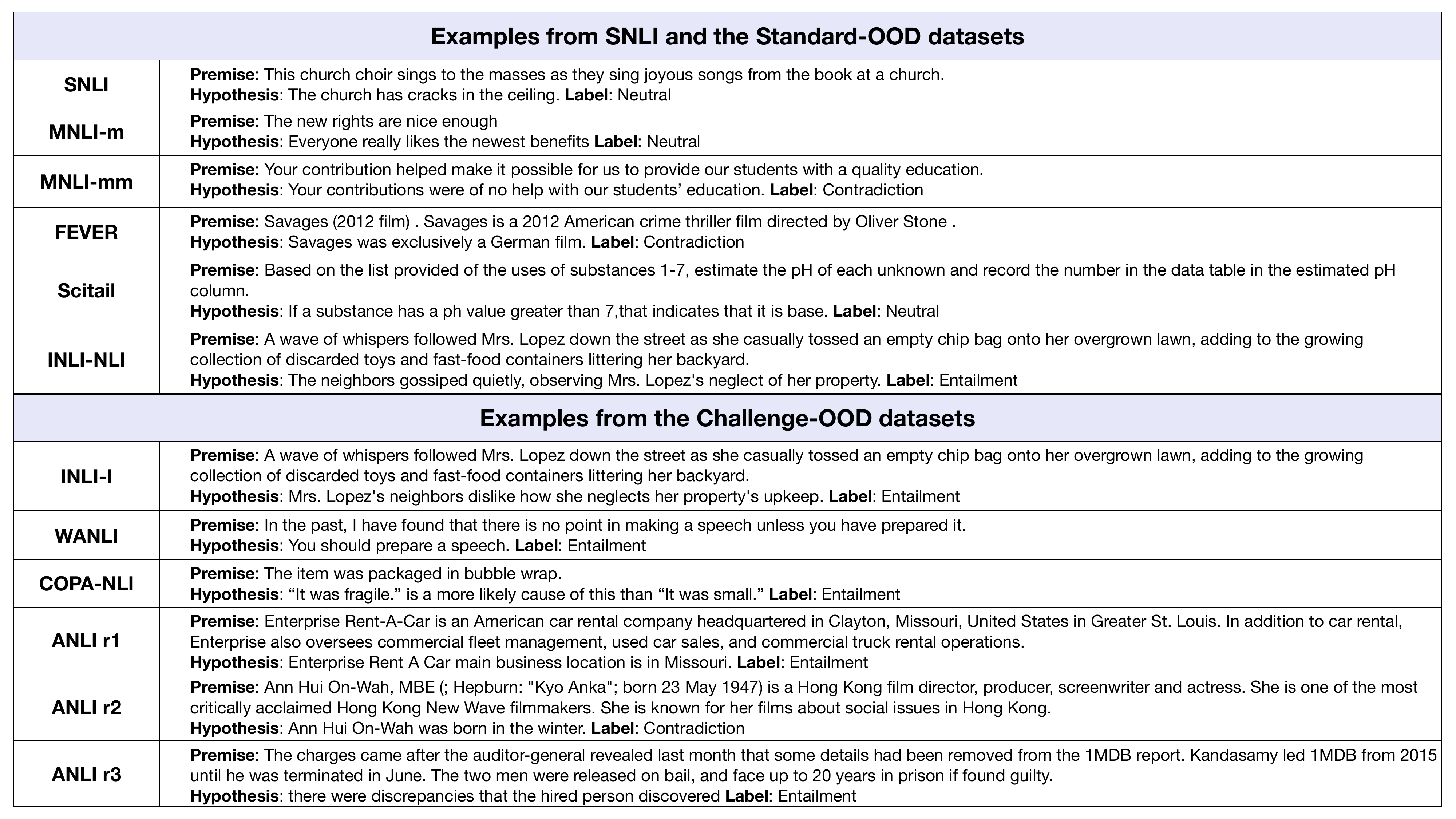} 
    \caption{Examples from the different NLI test sets used for model evaluation. The examples from Challenge-OOD datasets are more difficult than those from SNLI or the Standard-OOD datasets.} 
    \label{appendix:dataset_examples}

\end{center}
\end{figure*}

\paragraph{Dataset examples.}
Figure \ref{appendix:dataset_examples} provides examples from each of the NLI test sets used in this work. 

\section{Costs for Alternative Data Augmentation Strategies}

Table \ref{tab:appendix_cost_of_other_methods} provides approximate costs for fine-tuning GPT-4o-mini with other data augmentation methods. These costs assume a fixed cost per instance, in line with the costs calculated from SNLI training examples. In contrast to other approaches, the methods introduced in this paper keep the number of training instances fixed without increasing the costs of fine-tuning.

\begin{table}
\centering
\resizebox{\columnwidth}{!}{
 \begin{tabular}{l c c}
  \toprule

Method & Size of augmented data & Approx. cost in USD  \\ 

  \midrule
Ours & 0 (maintains size) & 5 \\
WANLI\textsuperscript{1} & 103k & 55 \\
Human-CAD\textsuperscript{2} & 108k & 60 \\
DISCO\textsuperscript{3} & 165k & 90 \\
UnitedSynT5\textsuperscript{4} & 468k & 240 \\
Z-Aug\textsuperscript{5} & 593k & 300 \\
GNLI\textsuperscript{6} & 671k & 340 \\
IBADR\textsuperscript{7} & 1.1m & 550 \\
\bottomrule
  \end{tabular}}
\caption{Approximate costs for fine-tuning GPT-4o-mini with other data augmentation methods. We assume a set cost per training example in line with SNLI, where the augmented data is used with an initial training set of 10k instances. Costs are provided to the nearest 5 dollars. \textsuperscript{1}\citet{liu-etal-2022-wanli}, \textsuperscript{2}\citet{DBLP:conf/iclr/KaushikHL20}, \textsuperscript{3}\citet{chen-etal-2023-disco}, \textsuperscript{4}\citet{DBLP:journals/corr/abs-2412-09263}, \textsuperscript{5}\citet{wu-etal-2022-generating}, \textsuperscript{6}\citet{hosseini-etal-2024-synthetic}, \textsuperscript{7}\citet{wang-etal-2023-ibadr}.}
\label{tab:appendix_cost_of_other_methods}
\end{table}

% I may also need to mention about the SNLI data here.

%\section{Example Appendix}
%\label{sec:appendix}
%Appendix text
\end{document}